\documentclass[10pt,twocolumn,letterpaper]{article}

\usepackage{wacv}              

\usepackage{graphicx}
\usepackage{amsmath}
\usepackage{amssymb}
\usepackage{booktabs}

\usepackage{multirow}
\usepackage[normalem]{ulem}
\useunder{\uline}{\ul}{}
\usepackage{comment}
\usepackage{tikz}
\usepackage{pgfplots}
\usepackage{rotating}
\usepackage{amssymb}
\usepackage{pifont}
\newcommand{\cmark}{\ding{51}}%
\newcommand{\xmark}{\ding{55}}%
\pgfplotsset{compat=1.18}
\usepackage{makecell}

\usepackage[accsupp]{axessibility} 

\usepackage[pagebackref,breaklinks,colorlinks]{hyperref}

\usepackage[capitalize]{cleveref}
\crefname{section}{Sec.}{Secs.}
\Crefname{section}{Section}{Sections}
\Crefname{table}{Table}{Tables}
\crefname{table}{Tab.}{Tabs.}

\def\wacvPaperID{656} 
\def\confName{WACV}
\def\confYear{2025}

\begin{document}

\title{Exploiting Inter-Sample Information for Long-tailed Out-of-Distribution Detection}

\author{Nimeshika Udayangani,  Hadi M. Dolatabadi,  Sarah Erfani, and Christopher Leckie\\
The University of Melbourne\\
{\tt\small \{nhewadehigah@student., h.dolatabadi@, sarah.erfani@, caleckie@\}unimelb.edu.au}
}
\maketitle

\begin{abstract}
Detecting out-of-distribution~(OOD) data is essential for safe deployment of deep neural networks~(DNNs).
This problem becomes particularly challenging in the presence of long-tailed in-distribution~(ID) datasets, often leading to high false positive rates~(FPR) and low tail-class ID classification accuracy.
In this paper, we demonstrate that exploiting inter-sample relationships using a graph-based representation can significantly improve OOD detection in long-tailed recognition of vision datasets.
To this end, we use the feature space of a pre-trained model to initialize our graph structure. 
We account for the differences between the activation layer distribution of the pre-training vs.~training data, and actively introduce Gaussianization to alleviate any deviations from a standard normal distribution in the activation layers of the pre-trained model. 
We then refine this initial graph representation using graph convolutional networks~(GCNs) to arrive at a feature space suitable for long-tailed OOD detection.
This leads us to address the inferior performance observed in ID tail-classes within existing OOD detection methods. 
Experiments over three benchmarks CIFAR10-LT, CIFAR100-LT, and ImageNet-LT demonstrate that our method outperforms the state-of-the-art approaches by a large margin in terms of FPR and tail-class ID classification accuracy.
\end{abstract}

\section{Introduction}
\label{sec:intro}

\begin{figure*}
  \centering
    \begin{subfigure}{1.0\linewidth}
    \includegraphics[width=\textwidth]{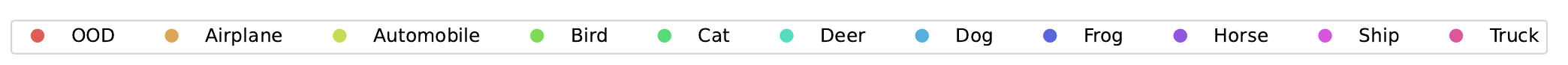}
    \caption*{}
    \label{fig:legend}
  \end{subfigure}
  \hfill
  \begin{subfigure}[t]{0.24\linewidth}
    \includegraphics[width=\textwidth]{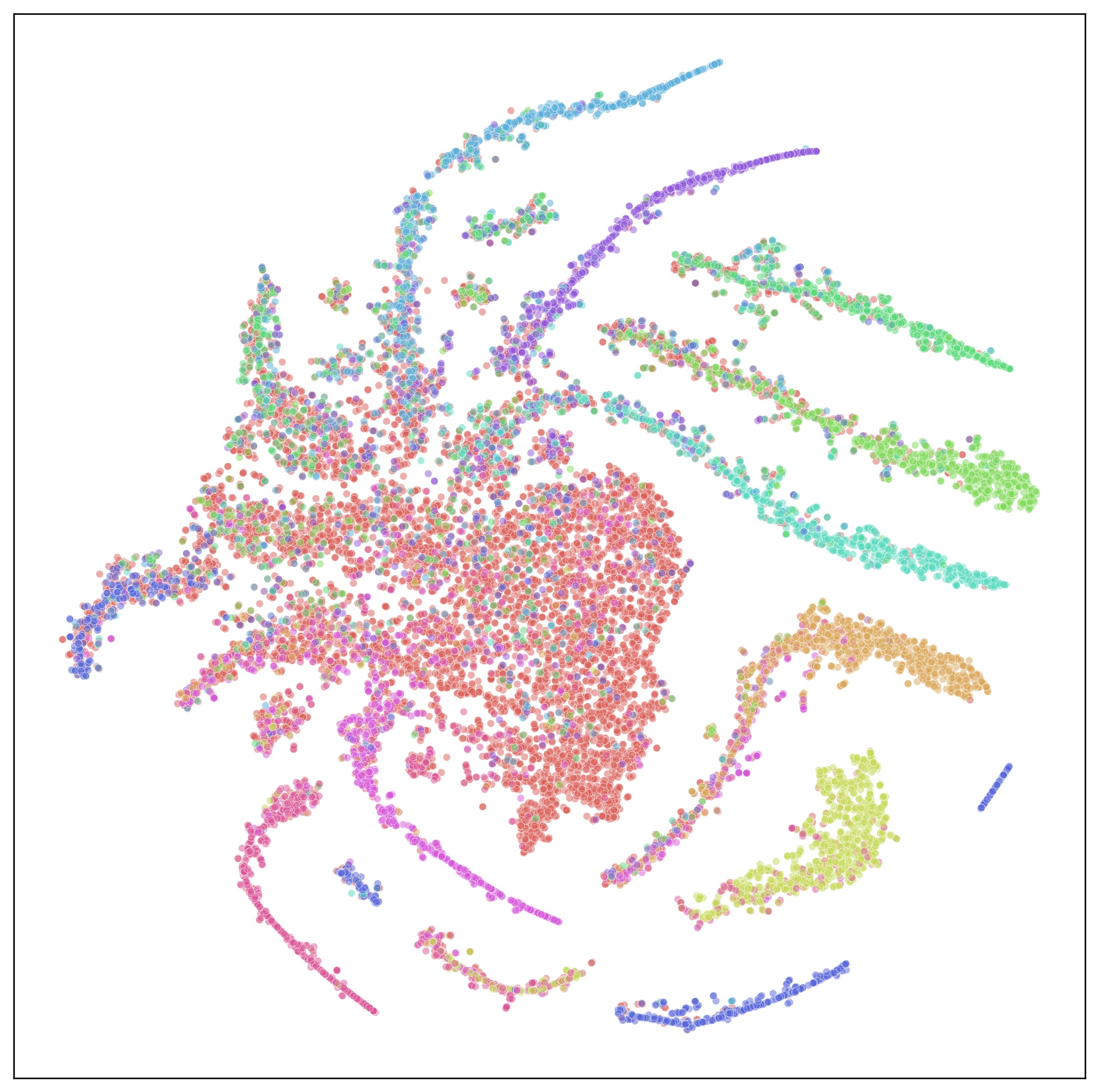}
    \caption{Scratch trained OE.}
    \label{fig:scratch-a}
  \end{subfigure}
  \hfill
  \begin{subfigure}[t]{0.24\linewidth}
    \includegraphics[width=\textwidth]{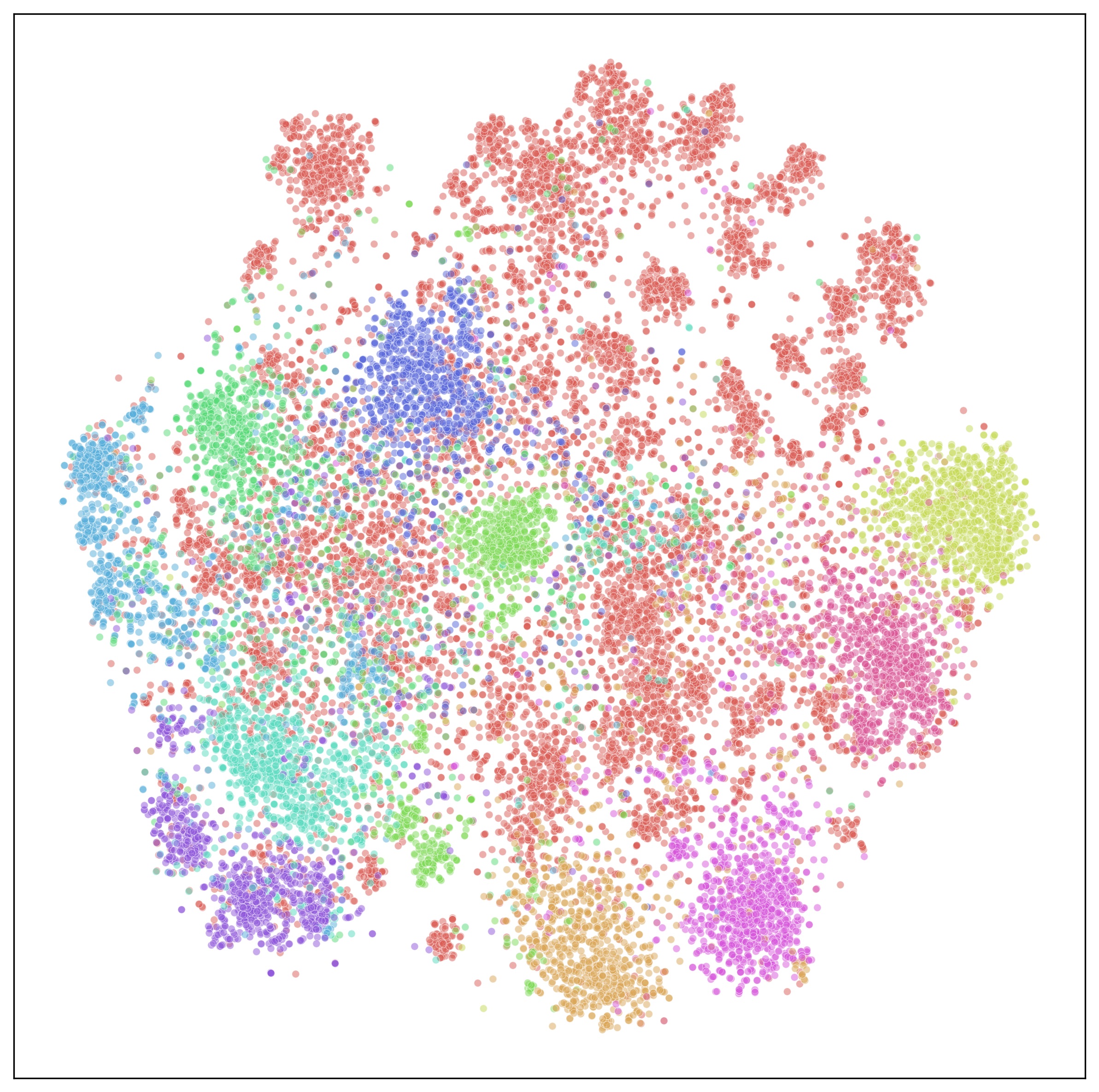}
    \caption{Pre-trained.}
    \label{fig:pretrain_only-b}
  \end{subfigure}
    \hfill
  \begin{subfigure}[t]{0.24\linewidth}
    \includegraphics[width=\textwidth]{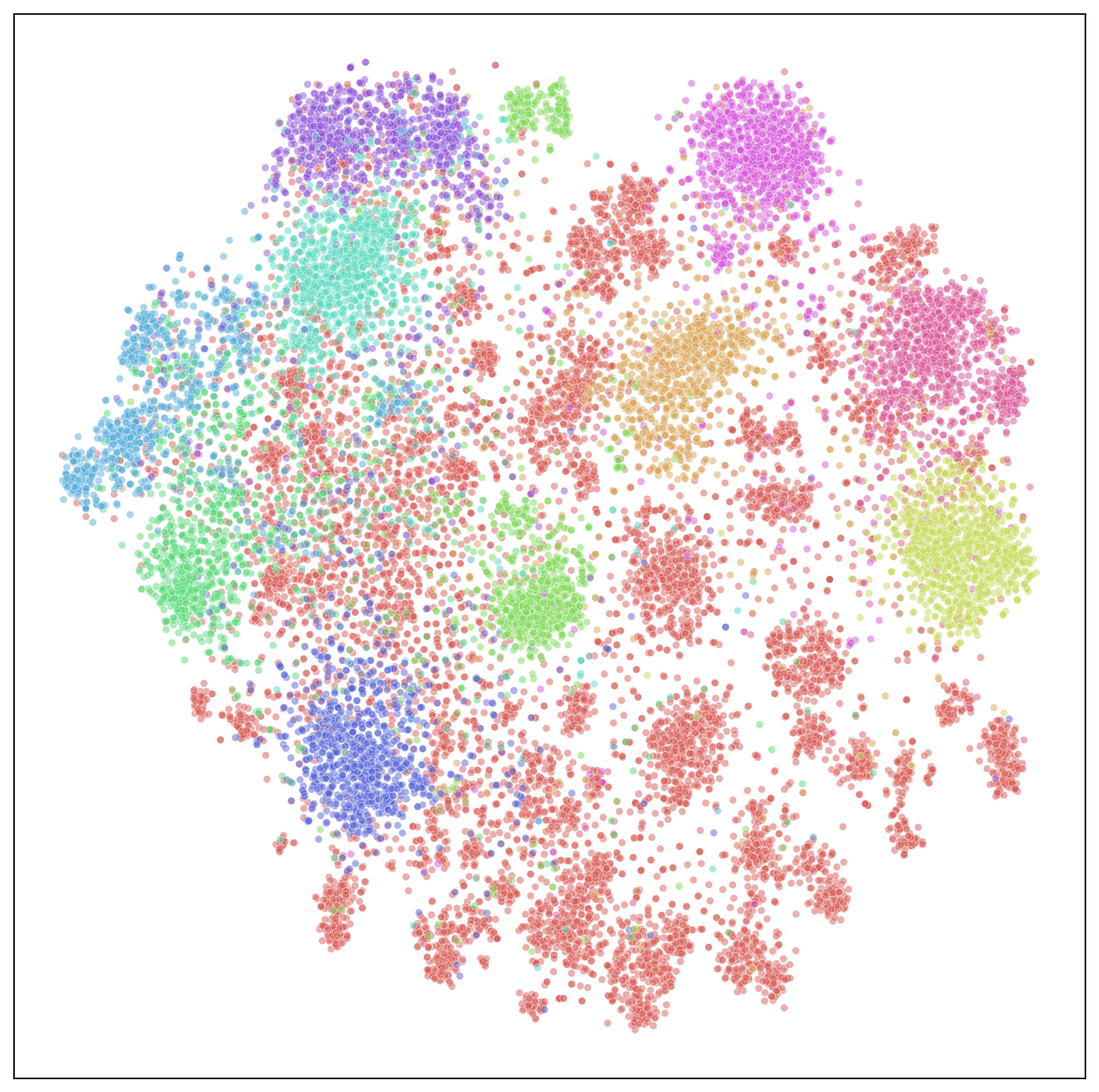}
    \caption{Pre-trained after Gaussianization.}
    \label{fig:pretrain_BN-c}
  \end{subfigure}
    \hfill
  \begin{subfigure}[t]{0.24\linewidth}
    \includegraphics[width=\textwidth]{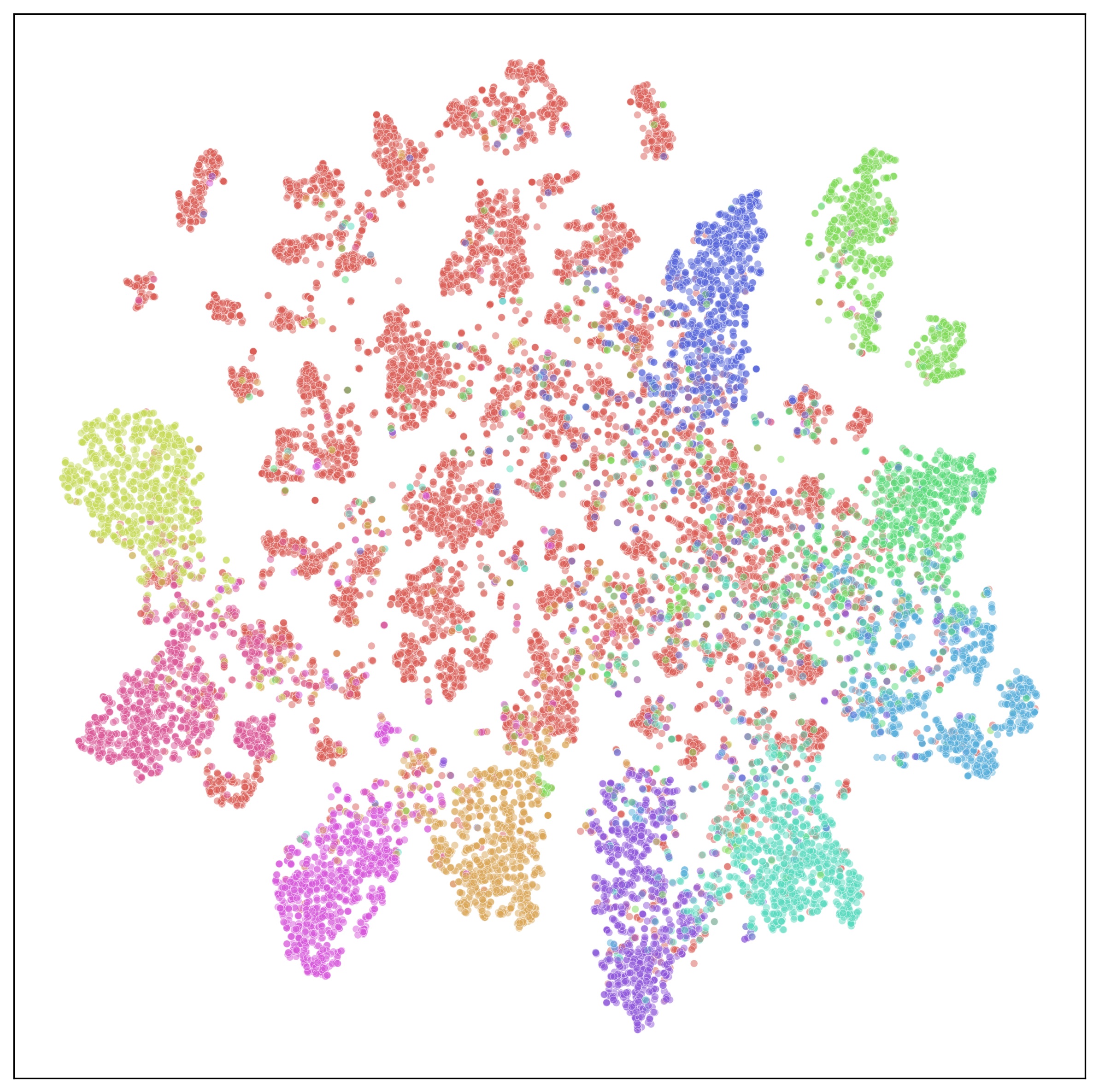}
    \caption{After GCN.}
    \label{fig:GCN-d}
  \end{subfigure}
  \caption{Feature space representations of CIFAR10-LT as ID test set and CIFAR-100 as OOD test set using t-SNE. (a) Feature distribution of ResNet18 trained from scratch using OE~\cite{hendrycks_2018OE}, (b) ResNet18 pre-trained on Downsampled ImageNet, (c) same model in (b) after applying Gaussianization, and (d) model in (c) after message passing using GCN. Tail ID classes (\eg, Horse, Ship, Truck) heavily overlap with OOD samples (represented as red dots) in (a). Separability enhanced from left to right, wherein (d) there are clear gaps between the decision boundaries. }
  \label{fig:feauture_space}
  \vspace{-0.15in}
\end{figure*}

Deep neural networks~(DNNs) exhibit remarkable performance provided that the data encountered during testing closely aligns with the data used for training. 
However, when DNN models encounter samples that are different from their training distribution~(a.k.a., out-of-distribution~(OOD) samples), they often assign overly confident predictions even when a distribution mismatch exists~\cite{DNN_fool}.
As a consequence, ensuring safe deployment of these models in critical domains, such as healthcare~\cite{health_ex} and autonomous driving~\cite{autonomous_ex}, necessitates methods that can reliably detect OOD samples. 

There have been numerous methods proposed to detect OOD samples showing promising results~\cite{hendrycks_2017MSP,hendrycks_2018OE,lee_2018_mahalanobis,EnergyOE,VIT_exploringlimitsOOD,KNN_OOD}.
However, most of these methods are commonly assessed using balanced class in-distributions~(IDs), which does not reflect real-world scenarios where the training data often exhibits a long-tailed~(LT) distribution. 
In a long-tailed distribution, a few dominant classes have a large number of samples~(head-classes), while the remaining classes~(tail-classes) have relatively fewer samples. 
As a result, models find it difficult to distinguish tail-class ID samples from OOD samples when compared to head-class ID samples. 
This leads to a high false positive rate~(FPR) for OOD detection and low tail-class ID classification accuracy in current methods, thereby making OOD detection extremely challenging~\cite{pascl}. 
It was not until very recently that Wang \etal~\cite{pascl} formally introduced the problem of OOD detection in long-tailed recognition~(LTR) for the first time. 
Following this trend, a few other methods~\cite{choiBalancedEnOE, EAT_AAAI, COCL_AAAI} evaluated their models on long-tailed training sets.

However, a common approach with these OOD detection methods --- as well as previous methods using balanced IDs --- is that they treat each sample individually during training, overlooking the interconnected relationships within the data.
We instead propose to exploit the inter-sample relationships within both ID and OOD data to address the more challenging problem of OOD detection in long-tailed settings. 
With the help of this additional relationship information, we aim to enhance the distinguishability of tail samples. 
To this end, we transform image datasets into graph-based representations, allowing us to utilize the full structure underlying the data.

In particular, we first use the feature space of a pre-trained model to initialize a graph structure using $k$ nearest neighbor~($k$-NN)~graphs. 
Even though there have been traces of pre-training being utilized to improve OOD detection~\cite{hendrycks_2019a_using_pre,VIT_exploringlimitsOOD,hendrycks_2020_pre_transformers}, none of these studies specifically explore the presence of long-tailed IDs. 
DNNs are unable to learn discriminative representations of tail-classes due to the scarcity of their training samples. 
Pre-training can alleviate this problem by exploiting prior knowledge from additional datasets, which can help to characterize the tail-classes. 
Due to the large-scale nature of these datasets, they often contain a variety of different samples, thus helping to achieve a more balanced performance when transferred to the long-tailed data. 
These ensure meaningful initial feature embeddings for our intended downstream task.
As shown for the CIFAR10-LT dataset in \cref{fig:feauture_space}, feature space representations from a pre-trained model (\cref{fig:pretrain_only-b}) facilitate a better separability among OOD and ID samples compared to a model trained from scratch (\cref{fig:scratch-a}) using Outlier Exposure~(OE)~\cite{hendrycks_2018OE}.

\looseness=-1
To further improve the quality of our initial representations, we account for the distributional differences between the activation layer of the pre-training vs.~training data. 
A recent study~\cite{jooBN} highlights the significance of ensuring that the activation layer representations of data adhere to a standard normal distribution.
To address this issue, we actively introduce Gaussianization to alleviate any deviations from a standard normal distribution in the activation layers of the pre-trained model during feature extraction~(\cref{fig:pretrain_BN-c}).
In this step, we update the batch norm~\cite{ioffe2015batchnorm} statistics of the pre-trained model to align it with our long-tailed distribution.
Finally, we refine this initial graph representation using message passing in the form of graph convolutional networks~(GCNs) to arrive at a feature space suitable for long-tailed OOD detection (\cref{fig:GCN-d}). 
To the best of our knowledge, we are the first to exploit inter-sample information 
to address the problem of long-tailed OOD detection in the vision domain. In summary, our contributions are as follows:
\begin{itemize}
\item 
We improve the tail-class representations utilizing inter-sample relationships by proposing a graph-based solution for OOD detection in long-tailed recognition.
\item We demonstrate that applying Gaussianization to the activation layers of the pre-trained model, using an auxiliary dataset to represent OOD data, can further improve the overall performance
\item Our method outperforms the SOTA by a large margin; (12.55\%, 8.51\%) FPR and (10.64\%, 12.95\%) tail-class ID classification accuracy on CIFAR10-LT and CIFAR100-LT, respectively. We also demonstrate our method can improve the performance on large-scale OOD detection tasks using ImageNet-LT.
\end{itemize}

\section{Related work}\label{sec:Related_work}
\looseness=-1
This section provides a brief overview of SOTA methods alongside baseline approaches, laying the groundwork for our research around OOD detection in LTR. 
We also delve into the significance of pre-training for OOD detection and explore the application of graph-based techniques for handling arbitrary data within our study.

\subsection{OOD detection}
\looseness=-1
Hendrycks and Gimpel~\cite{hendrycks_2017MSP} first suggested a baseline approach for identifying OOD samples using the maximum softmax probability~(MSP) from DNNs. 
This work revealed that OOD samples consistently exhibit lower MSP values compared to ID samples.
Similarly, to detect OOD samples other methods suggested different confidence scores (e.g., Mahalanobis distance~\cite{lee_2018_mahalanobis}, and Gram matrix values~\cite{gramMatrix_Sastry}) which can be applied to any pre-trained network. Very recently, KNN~\cite{KNN_OOD} suggested non-parametric nearest-neighbour distance for OOD detection.
While both our method and KNN employ $k$-NN, the usage is different. It utilizes $k$-NN as an OOD score, whereas we employ $k$-NN to establish our graph structure. Moreover, our method distinguishes itself from KNN by incorporating graph representations to leverage inter-sample relationships.
\looseness=-1

Another set of approaches~\cite{hendrycks_2018OE,EnergyOE,OECC_papadopo,CFL_by_r2} aim to leverage auxiliary datasets of outliers to improve the OOD detection. 
Rather than training solely on ID data, these approaches exposed DNNs to OOD examples. 
Outlier Exposure~(OE)~\cite{hendrycks_2018OE} modified the loss function of classification tasks by adding the KL divergence between the model's outputs for these training OOD samples and the uniform distribution. 
This modification led to improved performance compared to models trained exclusively on ID data. 
Following this method, EnergyOE~\cite{EnergyOE} and CFL~\cite{CFL_by_r2} used an energy score to differentiate between the auxiliary OOD samples and ID samples.  
Then, OECC~\cite{OECC_papadopo} proposed to minimize the total variation distance between the distributions instead of the KL divergence metric.
While these works propose promising solutions for OOD detection, their applicability is limited to datasets with a relatively balanced distribution. 

\subsection{Long-tailed OOD detection}
\looseness=-1
Despite the significant progress in OOD detection methods, very limited attention has been drawn to long-tailed IDs. 
PASCL~\cite{pascl} was the first to demonstrate the significant challenges of OOD detection in LTR and showed that this problem cannot be addressed by na\"ively combining SOTA OOD detection methods with long-tailed recognition methods. 
As a solution, PASCL applies supervised contrastive learning partially and asymmetrically to explicitly distinguish tail-class ID samples from OOD samples.
Recently, Balanced-EnergyOE~(BE-OE)~\cite{choiBalancedEnOE} updated the loss function of OE by introducing a balanced energy regularization loss, while OS~\cite{OS} assigned noisy labels to unlabeled OE data to re-balance the class priors in the training dataset.
However, these methods require fitting a prior distribution of OE data to ID data, which is difficult in LTR. To address this, COCL~\cite{COCL_AAAI} and EAT~\cite{EAT_AAAI} introduced abstention classes to better encapsulate the OOD samples and achieved SOTA results for OOD detection in LTR. 
Different from all these methods, our method is designed to exploit inter-sample information by means of graph representation learning to improve OOD detection.

\subsection{Pre-training for OOD detection}
\looseness=-1
Zeiler and Fergus~\cite{1st_ImageNet_pre_train} found that a pre-trained network on the ImageNet dataset \cite{ImageNet} can generalize well to other datasets, achieving a remarkable classification performance even with very limited numbers of training samples.  
Apart from classification tasks, Hendrycks \etal~\cite{hendrycks_2019a_using_pre} showed that pre-training can improve OOD detection on image datasets. 
They also demonstrated that certain properties, including the model's robustness to class imbalance, can be improved by pre-training. Similarly, pre-trained transformers have been shown to improve OOD detection in the domain of natural language processing~\cite{hendrycks_2020_pre_transformers}. 
Following this, Fort \etal~\cite{VIT_exploringlimitsOOD} showed that a pre-trained Vision Transformer~(ViT) can improve OOD detection in vision benchmarks. 
They also introduced few-shot OE to further improve OOD detection assuming that a few labelled samples from the test OOD dataset are available during training time.
Similarly, self-supervised pre-training has also been shown to improve OOD detection~\cite{hendrycks_2019b_using_self,ex2_self_sup_learning_forOOD}. 
However, none of these studies has examined how pre-training can improve OOD detection in the presence of long-tailed IDs. 
We are the first to incorporate pre-training to improve OOD detection in LTR.

\subsection{Graph neural networks for arbitrary data}
\looseness=-1
Following the concept of CNNs, GCNs~\cite{GCN_kipf} enable the generalization of neural network models to datasets structured as graphs. 
In terms of message passing~\cite{message_passing}, GCN can capture complex relationships and inter-dependencies among nodes in a graph. 
Since many modalities can be organized as graphs, GCNs made their way into the domain of computer vision \cite{zhang_global_localGCN,GCN_usage_face_cluster_1,GCN_usage_face_clustering_2,RobustImageClustering,CNN2Graph,misraaMultiModalRetrieval}. 
Zhang \etal~\cite{zhang_global_localGCN} deployed GCN for label noise cleansing in facial recognition tasks. 
Several other studies~\cite{GCN_usage_face_cluster_1,GCN_usage_face_clustering_2,RobustImageClustering} employed GCN for face image clustering.
The idea of GraphSAGE~\cite{GraphSAGE} was leveraged for tasks like image classification in CNN2Graph~\cite{CNN2Graph} and image retrieval in~\cite{misraaMultiModalRetrieval}. 
In this paper, however, we explore the introduction of GCN into the field of OOD detection in vision datasets. 
To the best of our knowledge, this is the first work that harnesses GCN to improve OOD detection alongside LTR.

\begin{figure*}[t]
  \centering
  \includegraphics[width=\textwidth]{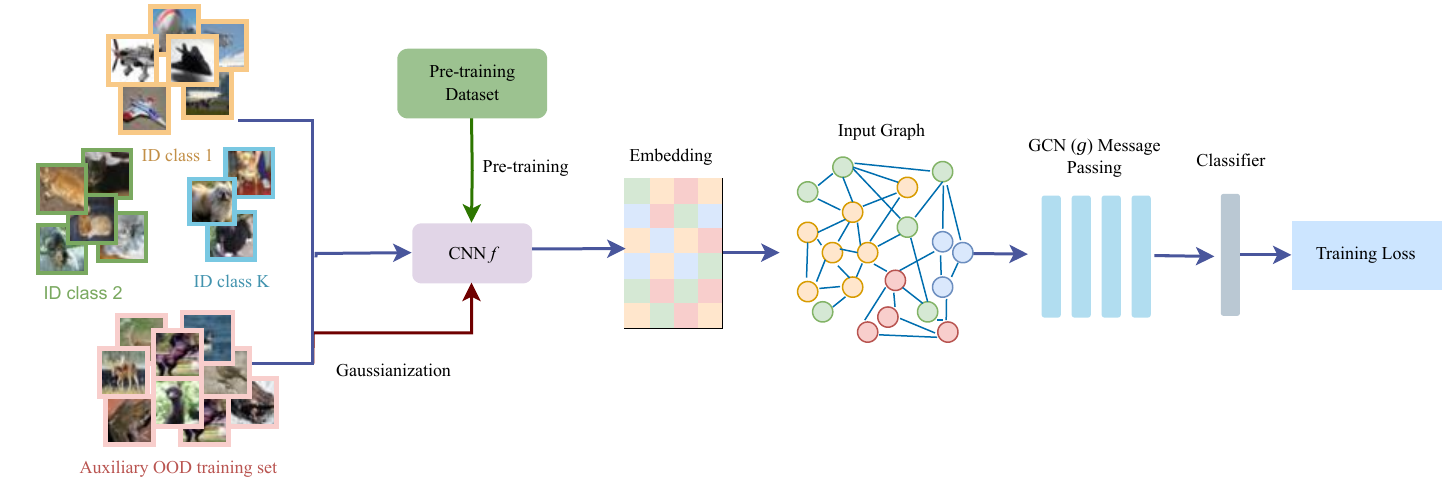}
   \vspace{-0.25in}
  \caption{Overview of proposed methodology. Given \(\mathcal{D}_{\mathrm{in}}\) with $K$ classes and and auxiliary OOD training set \(\mathcal{D}_{\mathrm{out}}^{\mathrm{OE}}\), we extract the initial feature embeddings from a pre-trained backbone model $f$ after applying Gaussianization using the same training data. We then create a $k$-NN graph to refine these initial features using message passing in the form of GCN $g$. Then, we incorporate a fully-connected classifier to arrive at the objective loss for training $g$. }
  \label{fig:pipeline}
  \vspace{-0.10in}
\end{figure*}

\section{Proposed method} \label{sec:method}
\looseness=-1
In this section, we propose our approach, which aims to improve the distinguishability of tail classes from OOD data by leveraging inter-sample relationships. To achieve this, we use graph representation learning to transform the input dataset into graph-based representations. We followed two important design choices to enhance the initial representations. 
First, we apply pre-training to obtain less biased feature representations. Next, we use Gaussianization to ensure that the activation layer representations conform to a standard normal distribution.
Finally, we refine this initial graph representation using GCN,  to arrive at a feature space suitable for long-tailed OOD detection.

We begin by providing a formal definition of our problem statement in \cref{sec:ProblemStatement}. 
Then, we describe each of the three major components that form the foundation of our methodology, as illustrated in~\cref{fig:pipeline}: 
(i) graph representation learning~(\cref{sec:GRL}), (ii) pre-training~(\cref{sec:preTraining}), and (iii) Gaussianization of pre-trained activation layers~(\cref{sec:BN}).

\subsection{Problem statement} \label{sec:ProblemStatement}
Let us consider a deep neural network model \(f\).
To train such models, we usually assume we have access to samples coming from a distribution referred to as \(\mathcal{D}_{\mathrm{in}}\) (\textbf{in-distribution}) (ID), consisting of \((\boldsymbol{x}_{\mathrm{in}}, y_{\mathrm{in}})\) pairs.
Here, \(\boldsymbol{x}_{\mathrm{in}}\) denotes the ID input image, and \( {y_{\mathrm{in}} \in \mathcal{Y}_{\mathrm{in}} :=\{1, ..., K\}}\) denotes its class label. 
The term \textbf{out-of-distribution}~(OOD) detection usually refers to identifying samples \(\boldsymbol{x}_{\mathrm{out}}\) that do not belong to \(\mathcal{D}_{\mathrm{in}}\) and their label set has no intersection with \(\mathcal{Y}_{\mathrm{in}}\). 
We denote this distribution as \(\mathcal{D}_{\mathrm{out}}\). 
In our case, the ID data \(\mathcal{D}_{\mathrm{in}}\) has a long-tailed distribution.
Without loss of generality, we assume that this distribution is characterized by an exponential decay in sample sizes across different classes with an imbalance ratio \(\rho\), which denotes the ratio between sample sizes of the most frequent and least frequent classes.
In other words, we assume \(\rho= \max_{i}\{n_{i}\}/ \min_{i}\{n_{i}\}\) where \(\left\{n_i\right\}_{i=1}^K\) represents the sample sizes for different classes.

In this problem setting, we aim to train neural networks $f$ such that they can: (i) correctly distinguish OOD samples $\boldsymbol{x}_{\mathrm{out}}$ from ID samples $\boldsymbol{x}_{\mathrm{in}}$, 
and (ii) classify ID samples into one of the \(K\) classes with a competitive accuracy for the less common ``tail-classes'' as those for the more prevalent ``head-classes''.
We can formulate our training objective as
  \vspace{-0.2in}
\begin{align}\nonumber
\mathcal{L} &= \mathbb{E}_{(\boldsymbol{x}_{\mathrm{in}},y_{\mathrm{in}})\sim \mathcal{D}_{\mathrm{in}}}\left[\mathcal{L}_{\mathrm{in}}(f(\boldsymbol{x}_{\mathrm{in}}),y_{\mathrm{in}})\right] \\ &\qquad\qquad+  \lambda \mathbb{E}_{\boldsymbol{x}_{\mathrm{out}} \sim \mathcal{D}_{\mathrm{out}}}\left[\mathcal{L}_{\mathrm{out}}\left(f\left(\boldsymbol{x}_{\mathrm{out}}\right)\right)\right] , \label{eq:training_obj}
\end{align}
where \(\mathcal{L}_{\mathrm{in}}\) is a loss function used for ID classification, and \(\mathcal{L}_{\mathrm{out}}\) aims to satisfy our objective over OOD data. 
Since we assume a realistic scenario where we have no prior knowledge of test-time OOD data during training, we use Outlier Exposure~(OE)~\cite{hendrycks_2018OE} and introduce an auxiliary dataset \(\mathcal{D}_{\mathrm{out}}^{\mathrm{OE}}\) that is entirely disjoint from test-time OOD data \(\mathcal{D}_{\mathrm{out}}^{\mathrm{test}}\). 
Using maximum softmax probability as the OOD score~\cite{hendrycks_2017MSP}, \(\mathcal{L}_{\mathrm{in}}\) in \cref{eq:training_obj} takes the form of cross-entropy loss, and \(\mathcal{L}_{\mathrm{out}}\) can be defined as the KL divergence between the model's output and the uniform distribution \(\mathcal{U}\): 
\begin{align}\nonumber
\mathcal{L} &= \mathbb{E}_{(\boldsymbol{x}_{\mathrm{in}},y_{\mathrm{in}})\sim \mathcal{D}_{\mathrm{in}}}\left[\mathcal{L}_{\mathrm{CE}}(f(\boldsymbol{x}_{\mathrm{in}}),y_{\mathrm{in}})\right] \\ &\qquad\qquad+ \lambda \mathbb{E}_{\boldsymbol{x}_{\mathrm{out}} \sim \mathcal{D}_{\mathrm{out}}^{\mathrm{OE}}}\left[\mathrm{KL}(f(\boldsymbol{x}_{\mathrm{out}}) \parallel {\mathcal{U}})\right].
\label{eq:vth_OE} 
\end{align}

\subsection{Graph representation learning} \label{sec:GRL}
\looseness=-1
As mentioned in \cref{sec:intro}, we want to have a training process that is fully aware of the overall structure underlying our dataset. 
Intuitively, we aim to ensure that the tail-class ID samples are not dominated by both the head-classes and the OOD samples.
To materialize this in the form of an inductive bias, we resort to organizing our input dataset using a graph structure. 
This way, we aim to improve the OOD detection in the presence of tail-classes by leveraging the inter-sample relationships within the data.
Existing OOD detection methods~\cite{hendrycks_2018OE,pascl,choiBalancedEnOE} only consider relationships between pairs or triplets of samples, or treat each sample completely individually to take the embedding space of the data. 
These methods, however, do not explore the embedding space in its entirety~\cite{GCN_usage_clusteringN_image_retreival}. 
To address this, we create a graph using all images in the training dataset to capture the relationships among all samples and apply message passing in the form of GCN~\cite{GCN_kipf}.

More precisely, we refine embeddings from a backbone CNN \(f\) using a GCN model \(g\) to obtain a more rectified embedding representation \(\boldsymbol{z}=g\left(f\left(\boldsymbol{x}\right)\right)\) before proceeding with the loss calculation in~\cref{eq:vth_OE}. 
Let \(\mathcal{X}\) denote the matrix of initial embeddings (the selection of initial embeddings is discussed further in the next section) from \(f\) for all samples in the training dataset (\(\left\{\mathcal{D}_{\mathrm {in}} \cup \mathcal{D}_{\mathrm{out}}^{\mathrm{OE}} \right\}\)). 
We calculate the pairwise cosine similarity \(S = \mathcal{XX}^{T}\) to build a \(k\)-nearest neighbors~(\(k\)-NN) undirected graph \(\mathcal{G}=(\mathcal{X},A)\) where \(A\) represents the adjacency matrix. 
If we consider the GCN operator from~\cite{GCN_kipf}, then the layer-wise forward propagation rule of \(g\) is given by:
\begin{equation} \label{eq:GCN_forward} 
H^{(l+1)}=\sigma\left(\tilde{D}^{-\frac{1}{2}} \tilde{A} \tilde{D}^{-\frac{1}{2}} H^{(l)} W^{(l)}\right),
\end{equation}
where $\tilde{A}=A+I_N$, and $I_N$ is the identity matrix. 
Also, we define $\tilde{D}_{i i}=\sum_j \tilde{A}_{i j}$ and $W^{(l)}$ is a trainable weight matrix in the $l^{\text {th }}$ layer. 
Finally, $H^{(l)}$ is the $l^{\text {th }}$ layer output matrix, with $H^{(0)}=\mathcal{X}$, and $\sigma$ denotes the activation function.

\looseness=-1
Instead of a CNN, which only considers intra-image local neighborhoods to calculate the feature representation of a sample, the GCN enables construction of an inter-image feature representation that aggregates neighboring information from multiple images. 
This additional relational information helps refine the poor representations of tail-class ID samples, as shown in \cref{fig:GCN-d}, resulting in a more significant margin between the decision boundary of tail-class ID and OOD samples.

It should be noted that we construct graphs in an inductive setting, where test data is not available to the training graph and the model can make inferences across different test graphs consisting of samples from \(\mathcal{D}_{\mathrm{in}}^{\mathrm{test}}\) and \(\mathcal{D}_{\mathrm{out}}^{\mathrm{test}}\).

\subsection{Pre-training} \label{sec:preTraining}
\looseness=-1
The selection of the initial feature embeddings vector \(\mathcal{X}\) is very important, as this will decide the overall structure underlying the input graph for the subsequent message passing and feature refinement process. 
One can choose a DNN model trained from scratch as the backbone network \(f\) above to extract the initial embeddings \(\mathcal{X}\). 
However, such an embedding distribution may be sub-optimal as the trained model is still biased towards the prevalent classes providing a poor prior for the downstream task.

It is well-known that pre-training is useful in leveraging the knowledge learned from a large, diverse dataset for specific tasks with limited data~\cite{1st_ImageNet_pre_train}. 
In our case, we have very limited tail-class ID samples, thus, it is conceivable that pre-training can be used to improve their representations. 
\Cref{fig:feauture_space} illustrates this fact by showing improved discrimination between the tail classes and OOD samples in the initial feature embedding space of a pre-trained model (\cref{fig:pretrain_only-b}) compared to a model trained from scratch (\cref{fig:scratch-a}). Therefore, we use a model pre-trained on a large dataset that is independent from \(\mathcal{D}_{\mathrm{in}}\), \(\mathcal{D}_{\mathrm{out}}^{\mathrm{OE}}\), and \(\mathcal{D}_{\mathrm{out}}^{\mathrm{test}}\) to generate \(\mathcal{X}\).

\subsection{Gaussianization of pre-trained activation layers} \label{sec:BN}
Regularization and normalization techniques are commonly applied to neural networks. 
A recent study~\cite{jooBN} highlights the significance of ensuring that the activation layer representations of data adhere to a standard normal distribution. 
However, when we forward pass our training datasets through a DNN that has been pre-trained on a different dataset, this can lead to a misalignment of the activation layers from the intended standard normal distribution. 
To rectify this misalignment, we apply Gaussianization to adjust the sample mean and sample variance of each activation layer of the pre-trained network to match those of the new training data, ensuring that their activations conform to a standard normal distribution. Accordingly, before extracting \(\mathcal{X}\) from the pre-trained model \(f\), we update its batch normalization~(BN)~\cite{ioffe2015batchnorm} statistics by fine-tuning it for an additional set of epochs on all the training data. 
Feature space representation after this process~(\cref{fig:pretrain_BN-c}) is substantially improved compared to the original one~(\cref{fig:pretrain_only-b}).


\section{Evaluation}\label{sec:Evaluation}
In this section, we evaluate the performance of our proposed model in terms of (i) OOD detection, and (ii) ID classification. 
We describe the experimental settings that we used. 
Our results demonstrate the effectiveness of our approach in comparison with the baselines, achieving SOTA performance in all settings.
Additionally, we conduct thorough ablation studies to demonstrate the effectiveness of each component of our method.

\subsection{Experimental settings} \label{sec:experimental_setup}

\noindent\textbf{ID datasets~(\(\mathcal{D}_{\mathrm{in}}\) , \(\mathcal{D}_{\mathrm{in}}^{\mathrm{test}}\)):}
 We use three long-tailed datasets, CIFAR10-LT, CIFAR100-LT~\cite{cifarLT} with the imbalance ratio $\rho=100$ following~\cite{pascl,EAT_AAAI,COCL_AAAI}, and ImageNet-LT~\cite{OLTR_or_ImageNetLT} as ID data. 
 As \(\mathcal{D}_{\mathrm{in}}^{\mathrm{test}}\), we use balanced test sets of the original CIFAR datasets~\cite{CIFAR_normal} and the validation set of ImageNet~\cite{ImageNet}.

\noindent\textbf{OE datasets~(\(\mathcal{D}_{\mathrm{out}}^{\mathrm{OE}}\)):} We use two auxiliary datasets, 80 Million Tiny Images~\cite{torralba80MillionTiny2008} in the case of training over CIFAR10/100-LT, and ImageNet-Extra\footnote{\label{ft:pascl}https://github.com/amazon-science/long-tailed-ood-detection} provided in~\cite{pascl} when ID is ImageNet-LT.

\noindent\textbf{OOD datasets~(\(\mathcal{D}_{\mathrm{out}}^{\mathrm{test}}\)):} For the OOD test data we use six datasets---Texture~\cite{textures}, SVHN~\cite{SVHN}, LSUN~\cite{lsun}, Places365~\cite{places365}, CIFAR, and TinyImagenet~\cite{tinyImageNet}--- provided in the SC-OOD benchmark~\cite{SCOOD_bench} when ID is CIFAR10/100-LT. ImageNet-1k-OOD\footref{ft:pascl} was used as \(\mathcal{D}_{\mathrm{out}}^{\mathrm{test}}\) when ID is ImageNet-LT following~\cite{pascl, EAT_AAAI, COCL_AAAI}.

\noindent\textbf{Evaluation metrics:} We use six different metrics following existing  methods~\cite{pascl,choiBalancedEnOE,EAT_AAAI} to compare the effectiveness of our model: \textbf{(1) FPR95} measures the false positive rate of OOD samples when the true positive rate of ID samples is 95\%; \textbf{(2) AUROC} is the area under the receiver operating characteristic
curve; and \textbf{(3) AUPR} is the area under the precision-recall curve. To measure the ID classification performance we use: \textbf{(4) ACC} as the accuracy on the entire ID test set; \textbf{(5) ACC\textsubscript{head}} represents the accuracy of the first 50\% of the most redundant ID classes; and \textbf{(6) ACC\textsubscript{tail}}, the accuracy of the rest of the classes.

\noindent\textbf{Pre-trained models:} As the backbone pre-trained model \(f\), we use the same model architectures used in \cite{pascl,choiBalancedEnOE,EAT_AAAI} for fair comparisons. For experiments on CIFAR10/100-LT, we use the ResNet18~\cite{ResNet} model pre-trained using \(32\times32\) resolution Downsampled ImageNet~\cite{downsampled_ImageNet} dataset following~\cite{hendrycks_2019a_using_pre}. For experiments on ImageNet-LT, the self supervised pre-training model DINO~\cite{DINO} with ResNet50~\cite{ResNet} architecture is used.
We carefully select these models to ensure there is no data leakage between the model weights and the data from \(\mathcal{D}_{\mathrm{in}}\) or \(\mathcal{D}_{\mathrm{out}}\).

\noindent\textbf{Graph models:} A 3-layer GCN with 512-dimensional hidden features is used as \(g\) when ID is CIFAR10/100-LT. For ImageNet-LT, the hidden dimension is increased to 2048.

\begin{table}[t!]
\centering
\caption{OOD detection results for CIFAR10-LT as ID. The best results are shown in bold. Mean over six random runs are reported.}
\vspace{-0.10in}
\begin{footnotesize}
\setlength{\tabcolsep}{7pt} 
\renewcommand{\arraystretch}{0.85} 
\begin{tabular}{c|c|c|c|c}
\hline
\textbf{$\mathcal{D}_{\mathrm{out}}^{\mathrm{test}}$}           & \textbf{Method}   & \textbf{AUROC }& \textbf{AUPR}  & \textbf{FPR95} \\ \hline
\multirow{4}{*}{\rotatebox[origin=c]{90}{Texture}}      
                              & OE                & 92.59          & 83.32          & 25.10          \\
                              & PASCL             & 93.16          & 84.80          & 23.26          \\
                              & BE-OE             & 95.69     & 92.38          & 21.26          \\
 & EAT& 95.44& 92.28&21.50\\
                              & Ours              & \textbf{99.50} & \textbf{98.92} & \textbf{1.10}  \\ \hline
\multirow{4}{*}{\rotatebox[origin=c]{90}{SVHN}}         
                              & OE                & 95.10          & 97.14          & 16.15          \\
                              & PASCL             & 96.63          & 98.06          & 12.18          \\
                              & BE-OE             & 97.74          & 98.89          & 9.87           \\
 & EAT& 97.92& 99.06&9.87\\
                              & Ours              & \textbf{99.68} & \textbf{99.71} & \textbf{0.67}  \\ \hline
\multirow{4}{*}{\rotatebox[origin=c]{90}{CIFAR100}}     
                              & OE                & 83.40          & 80.93          & 56.96          \\
                              & PASCL             & 84.43          & 82.99          & 57.27          \\
                              & BE-OE             & 85.20          & 84.98          & 57.95          \\
 & EAT& 85.93& 86.10&54.13\\
                              & Ours              & \textbf{92.08} & \textbf{92.21} & \textbf{35.28} \\ \hline
\multirow{4}{*}{\rotatebox[origin=c]{90}{\makecell{Tiny-\\ImageNet}}} 
                              & OE                & 86.14          & 79.33          & 47.78          \\
                              & PASCL             & 87.14          & 81.54          & 47.69          \\
                              & BE-OE             & 88.92          & 84.98          & 42.38          \\
 & EAT& 89.11& 85.43&41.75\\
                              & Ours              & \textbf{95.61} & \textbf{93.64} & \textbf{19.93} \\ \hline
\multirow{4}{*}{\rotatebox[origin=c]{90}{LSUN}}         
                              & OE                & 91.35          & 87.62          & 27.86          \\
                              & PASCL             & 93.17          & 91.76          & 26.40          \\
                              & BE-OE             & 94.48          & 93.15          & 23.88          \\
 & EAT& 95.13& 94.12&19.72\\
                              & Ours              & \textbf{99.57} & \textbf{99.43} & \textbf{0.74}  \\ \hline
\multirow{4}{*}{\rotatebox[origin=c]{90}{Places365}}    
                              & OE                & 90.07          & 95.15          & 34.04          \\
                              & PASCL             & 91.43          & 96.28          & 33.40          \\
                              & BE-OE             & 93.35          & 97.23          & 28.25          \\
 & EAT& 93.68& 97.42&26.03\\
                              & Ours              & \textbf{98.12} & \textbf{99.12} & \textbf{5.89}  \\ \hline
\multirow{4}{*}{\rotatebox[origin=c]{90}{Average}}      
                              & OE                & 89.77          & 87.25          & 34.65          \\
                              & PASCL             & 90.99          & 89.24          & 33.36          \\
                              & BE-OE             & 92.56          & 91.94          & 30.60          \\
 & EAT& 92.87& 92.40&28.83\\
                              & Ours              & \textbf{97.43} & \textbf{97.17} & \textbf{10.60} \\ \hline
\end{tabular}
\end{footnotesize}
\label{tab:cifar10_results}
\vspace{-0.20in}
\end{table}

\begin{table}[t!]
\centering
\caption{OOD detection results for CIFAR100-LT as ID. The best results are shown in bold. Mean over six random runs are reported.}
\vspace{-0.10in}
\begin{footnotesize}
\renewcommand{\arraystretch}{0.85} 
\begin{tabular}{c|c|c|c|c}
\hline
\textbf{$\mathcal{D}_{\mathrm{out}}^{\mathrm{test}}$}           & \textbf{Method}   & \textbf{AUROC} & \textbf{AUPR}  & \textbf{FPR95} \\ \hline
\multirow{4}{*}{\rotatebox[origin=c]{90}{Texture}}      
                              & OE                & 76.71          & 58.79          & 68.28          \\
                              & PASCL             & 76.01          & 58.12          & 67.43          \\
                              & BE-OE & 82.10          & 73.09          & 64.19          \\
 & EAT& 80.27& 71.76&67.53\\
                              & Ours              & \textbf{90.88} & \textbf{84.10} & \textbf{39.32} \\ \hline
\multirow{4}{*}{\rotatebox[origin=c]{90}{SVHN}}         
                              & OE                & 77.61          & 86.82          & 58.04          \\
                              & PASCL             & 80.19          & 88.49          & 53.45          \\
                              & BE-OE & 88.66          & 92.88          & 33.79          \\
 & EAT& 83.11& 89.71&47.78\\
                              & Ours              & \textbf{97.02} & \textbf{98.19} & \textbf{10.19} \\ \hline
\multirow{4}{*}{\rotatebox[origin=c]{90}{CIFAR10}}      
                              & OE                & 62.23          & 57.57          & 80.64          \\
                              & PASCL             & 62.33          & 57.14          & 79.55          \\
                              & BE-OE & 59.40          & 54.97          & 85.16          \\
 & EAT& 61.62& 55.30&77.97\\
                              & Ours              & \textbf{75.08} & \textbf{74.17} & \textbf{70.62} \\ \hline
\multirow{4}{*}{\rotatebox[origin=c]{90}{\makecell{Tiny-\\ImageNet}}}
                              & OE                & 68.04          & 51.66          & 76.66          \\
                              & PASCL             & 68.20          & 51.53          & 76.11          \\
                              & BE-OE & 71.42          & 56.52          & 74.22          \\
 & EAT& 68.34& 52.79&74.89\\
                              & Ours              & \textbf{80.06} & \textbf{68.96} & \textbf{63.48} \\ \hline
\multirow{4}{*}{\rotatebox[origin=c]{90}{LSUN}}         
                              & OE                & 77.10          & 61.42          & 63.98          \\
                              & PASCL             & 77.19          & 61.27          & 63.31          \\
                              & BE-OE & \textbf{83.83} & 71.23          & \textbf{52.04} \\
 & EAT& 81.09& 67.46&55.02\\
                              & Ours              & 83.29          & \textbf{75.09} & 59.57          \\ \hline
\multirow{4}{*}{\rotatebox[origin=c]{90}{Places365}}    
                              & OE                & 75.80          & 86.68          & 65.72          \\
                              & PASCL             & 76.02          & 86.52          & 64.81          \\
                              & BE-OE & 81.10          & 89.94          & \textbf{57.52} \\
 & EAT& 78.28& 88.20&60.85\\
                              & Ours              & \textbf{84.22} & \textbf{92.63} & 58.87          \\ \hline
\multirow{4}{*}{\rotatebox[origin=c]{90}{Average}}      
                              & OE                & 72.91          & 67.16          & 68.89          \\
                              & PASCL             & 73.32          & 67.18          & 67.44          \\
                              & BE-OE & 77.75          & 73.10          & 61.15          \\
 & EAT& 75.45& 70.87&64.01\\ 
                              & Ours              & \textbf{85.09} & \textbf{82.19} & \textbf{50.34} \\ \hline
\end{tabular}
\end{footnotesize}
\label{tab:cifar100_results}
\vspace{-0.20in}
\end{table}

\noindent\textbf{Baseline methods:} We compare our method with SOTA OOD detection methods in LTR: PASCL~\cite{pascl}, Balanced EnergyOE~(BE-OE)~\cite{choiBalancedEnOE}, OS~\cite{OS}, COCL~\cite{COCL_AAAI} and EAT~\cite{EAT_AAAI}. Three other popular OOD detection methods --- OE~\cite{hendrycks_2018OE}, EnergyOE~\cite{EnergyOE}, and OECC~\cite{OECC_papadopo} --- are also compared. For methods using pre-training for OOD detection, we include Pre-train+MSP~\cite{hendrycks_2019a_using_pre} and Pre-train+Mahalan.~\cite{VIT_exploringlimitsOOD}. These methods employ the same pre-trained models utilized in our experiments, which are subsequently fine-tuned on training data. We also compare the results with KNN~\cite{KNN_OOD}, which also finds $k$-nearest neighbours for OOD detection, but does not utilize graph representation learning unlike in our method. For a fair comparison, we incorporate a fine-tuned pre-trained model to apply the $k$-NN distance (Pre-train+KNN). All these methods use auxiliary OOD data for a optimum performance and the same model architectures for fair comparisons.
More details on hyper-parameter tuning can be found in the supplementary material.

\begin{table*}[tb!]
\centering
\caption{OOD detection results and ID classification results for CIFAR10-LT, CIFAR100-LT and ImageNet-LT as ID, with additional baselines. For CIFAR10-LT and CIFAR100-LT average values on all six \(\mathcal{D}_{\mathrm{out}}^{\mathrm{test}}\) are reported. The best and second-best results are bolded and underlined, respectively. Mean over six random runs are reported.}
\begin{footnotesize}
\renewcommand{\arraystretch}{1.01} 
\begin{tabular}{c|c|c|ccc|ccc}
\hline
$\mathcal{D}_{\mathrm{in}}$   & $\mathcal{D}_{\mathrm{out}}^{\mathrm{test}}$  & \textbf{Method} & \textbf{AUROC} & \textbf{AUPR}  & \textbf{FPR95} & \textbf{ACC}   & \textbf{ACC}\textsubscript{\textbf{head}} & \textbf{ACC}\textsubscript{\textbf{tail}} \\ \hline
\multirow{9}{*}{\rotatebox[origin=c]{90}{CIFAR10-LT}}        & \multirow{9}{*}{SC-OOD}                  & OECC                                 & 87.28                               & 86.29                              & 45.24                               & 61.24                             & 84.96                                  & 37.52                                  \\
                                   &                                          & EnergyOE                             & 91.92                               & 91.97                              & 33.79                               & 74.25                             & 87.74                                  & 60.76                                  \\
                                   &                                          & OE                                   & 89.77                               & 87.25                              & 34.65                               & 73.86                             & 84.92                                  & 62.80                                  \\
 & & OS& 91.94& 89.35& 36.92& 75.61& 84.32&66.90\\
                                   &                                          & PASCL                                & 90.99                               & 89.24                              & 33.36                               & 77.07                             & 86.12                                  & 68.02                                  \\
                                   &                                          & BE-OE                                & 92.56                               & 91.94                              & 30.60                               & 81.32                             & 85.14                                  & 77.50                                  \\
 & & EAT& 92.87& 92.40& 28.83& 81.42& 85.94&76.90\\
 & & COCL& 93.28& 92.89& 30.88& 81.56& 85.22&77.90\\
                                   &                                          & Pre-train+KNN                        & 90.20                               & 89.59                              & 55.53                               & {\ul 85.66}                       & \textbf{93.20}                         & {\ul 78.12}                            \\
                                   &                                          & Pre-train+MSP                        & {\ul 94.10}                         & 92.91                              & {\ul 23.15}                         & {\ul 85.66}                       & \textbf{93.20}                         & {\ul 78.12}                            \\
                                   &                                          & Pre-train+Mahalan.                   & 92.89                               & {\ul 93.01}                        & 28.79                               & {\ul 85.66}                       & \textbf{93.20}                         & {\ul 78.12}                            \\
                                   &                                          & Ours                                 & \textbf{97.43}                      & \textbf{97.17}                     & \textbf{10.60}                      & \textbf{89.74}                    & {\ul 90.71}                            & \textbf{88.76}                         \\ \hline
\multirow{9}{*}{\rotatebox[origin=c]{90}{CIFAR100-LT}}      & \multirow{9}{*}{SC-OOD}                  & OECC                                 & 70.38                               & 66.87                              & 73.15                               & 32.93                             & 52.31                                  & 13.02                                  \\
                                   &                                          & EnergyOE                             & 76.40                               & 72.24                              & 64.54                               & 40.63                             & 63.30                                  & 17.96                                  \\
                                   &                                          & OE                                   & 72.91                               & 67.16                              & 68.89                               & 39.51                             & 60.98                                  & 18.04                                  \\
 & & OS& 74.37& 70.42& 78.18& 40.92& 58.49&23.34\\
                                   &                                          & PASCL                                & 73.32                               & 67.18                              & 67.44                               & 43.09                             & 59.50                                  & 26.68                                  \\
                                   &                                          & BE-OE                                & 77.75                               & 73.10                              & 61.15                               & 45.72                             & 56.72                                  & 34.72                                  \\
 & & EAT& 75.45& 70.87& 64.01& 45.31& 62.76&27.86\\
 & & COCL& 78.25& 73.58& 74.09& 46.41& 63.38&29.44\\
                                   &                                          & Pre-train+KNN                        & 72.57                               & 67.28                              & 77.03                               & {\ul 56.02}                       & {\ul 73.82}                            & {\ul 38.22}                            \\
                                   &                                          & Pre-train+MSP                        & {\ul 79.04}                         & {\ul 74.00}                        & {\ul 58.85}                         & {\ul 56.02}                       & {\ul 73.82}                            & {\ul 38.22}                            \\
                                   &                                          & Pre-train+Mahalan.                   & 74.91                               & 72.38                              & 67.30                               & {\ul 56.02}                       & {\ul 73.82}                            & {\ul 38.22}                            \\
                                   &                                          & Ours                                 & \textbf{85.09}                      & \textbf{82.19}                     & \textbf{50.34}                      & \textbf{62.97}                    & \textbf{74.76}                         & \textbf{51.17}                         \\ \hline
\multirow{9}{*}{\rotatebox[origin=c]{90}{ImageNet-LT}}       & \multirow{9}{*}{\makecell{ImageNet-\\1k-OOD}}        & OECC                                 & 63.07                               & 63.05                              & 86.90                               & 38.25                             & 54.45                                  & 22.71                                  \\
                                   &                                          & EnergyOE                             & 65.33                               & 67.40                              & 88.22                               & 43.27                             & 62.81                                  & 23.72                                  \\
                                   &                                          & OE                                   & 66.33                               & 68.29                              & 88.22                               & 37.60                             & 54.29                                  & 20.90                                  \\
 & & OS& 68.01& 69.53& 88.97& 45.23& 56.25&34.21\\
                                   &                                          & PASCL                                & 68.00                               & 70.15                              & 87.53                               & 45.49                             & 54.73                                  & 36.26                                  \\
                                   &                                          & BE-OE                                & 66.89                               & 68.81                              & 87.70                               & 47.81                       &58.31                           & {\ul 37.32}                            \\
 & & EAT& 69.84& 69.25& 87.63& 46.79& 59.46&34.12\\
 & & COCL& 69.93& 69.78& 87.46& {\ul 51.11}& \textbf{74.17}&28.05\\
                                   &                                          & Pre-train+KNN                        & 70.65                         & 70.42                              & \textbf{78.14}                      & 48.79                             & {\ul68.02}                                  & 29.56                                  \\
                                   &                                          & Pre-train+MSP                        & {\ul 72.27}                      & {\ul 74.46}                     & 85.09                               & 48.79                             & {\ul68.02}                                   & 29.56                                  \\
                                   &                                          & Pre-train+Mahalan.                   & 68.91                               & 69.84                              & 85.91                         & 48.79                             & {\ul68.02}                                   & 29.56                                  \\
                                   &                                          & Ours                                 & \textbf{74.76}                               & \textbf{75.99}                              & {\ul 82.42}                        & \textbf{51.57}                    & 64.84                         & \textbf{38.31}                         \\ \hline
\end{tabular}    
\end{footnotesize}
\label{tab:main_results}
\vspace{-0.05in}
\end{table*}

\begin{table*}[tb!]
\centering
\caption{Ablation study. Average values on all six \(\mathcal{D}_{\mathrm{out}}^{\mathrm{test}}\) are reported. The best and second-best results are bolded and underlined, respectively. Mean over six random runs are reported.}
\begin{footnotesize}
\renewcommand{\arraystretch}{1.01} 
\begin{tabular}{c|c|c|c|c|c|c|c|c|c|c}
\hline
$\mathcal{D}_{\mathrm{in}}$                 & \textbf{\makecell{Pre-train}} & {\textbf{Gau.}} & \textbf{GRL} & \textbf{Method}              & \textbf{AUROC} & \textbf{AUPR}  & \textbf{FPR95} & \textbf{ACC}   & \textbf{ACC}\textsubscript{\textbf{head}} & \textbf{ACC}\textsubscript{\textbf{tail}}\\ \hline
\multirow{6}{*}{\rotatebox[origin=c]{90}{CIFAR10-LT}}  & \xmark                     & \xmark               & \xmark            & Scratch (OE)                 & 89.77          & 87.25          & 34.65          & 73.86&   84.92&    62.80\\
                             & \xmark                     & \xmark               & \cmark            & Scratch+GCN                  & 84.25          & 79.20          & 39.22          & 66.73          & 76.84             & 56.62             \\
                             & \cmark                     & \xmark               & \xmark            & Pre-train                     & 91.37          & 91.29          & 36.36          & 76.34          & {\ul 92.92}       & 59.76             \\
                             & \cmark                     & \xmark               & \cmark            & Pre-train+GCN                 & {\ul 96.33}    & {\ul 95.89}    & {\ul 14.28}    & {\ul 86.69}    & 90.15             & {\ul 83.22}       \\
                             & \cmark                     & \cmark               & \xmark            & Pre-train+Gau.              & 92.59          & 93.15          & 34.06          & 80.23          & \textbf{94.18}    & 66.28             \\
                             & \cmark                     & \cmark               & \cmark            & \makecell{Pre-train+Gau.+GCN(ours)} & \textbf{97.43} & \textbf{97.17} & \textbf{10.60} & \textbf{89.74} & 90.71             & \textbf{88.76}    \\ \hline
\multirow{6}{*}{\rotatebox[origin=c]{90}{CIFAR100-LT}} & \xmark                     & \xmark               & \xmark            & Scratch (OE)                 & 72.91          & 67.16          & 68.89          & 39.51          & 60.98             & 18.04             \\
                             & \xmark                     & \xmark               & \cmark            & Scratch+GCN                  & 73.93          & 65.97          & 68.58          & 40.74          & 57.66             & 23.82             \\
                             & \cmark                     & \xmark               & \xmark            & Pre-train                     & 72.17          & 68.43          & 71.23          & 50.38          & {\ul 74.92}       & 25.84             \\
                             & \cmark                     & \xmark               & \cmark            & Pre-train+GCN                 & {\ul 81.40}    & {\ul 78.05}    & {\ul 58.88}    & 54.48          & 69.45             & {\ul 39.51}       \\
                             & \cmark                     & \cmark               & \xmark            & Pre-train+Gau.              & 78.01          & 74.64          & 64.11          & {\ul 55.66}    & \textbf{77.94}    & 33.38             \\
                             & \cmark                     & \cmark               & \cmark            & \makecell{Pre-train+Gau.+GCN(ours)} & \textbf{85.09} & \textbf{82.19} & \textbf{50.34} & \textbf{62.97} & 74.76             & \textbf{51.17}    \\ \hline
\end{tabular}
\end{footnotesize}
\label{tab:ablation_results}
\vspace{-0.050in}
\end{table*}

\begin{figure*}[tb!]
  \centering
  \begin{subfigure}{0.32\linewidth}
    \includegraphics[width=\textwidth]{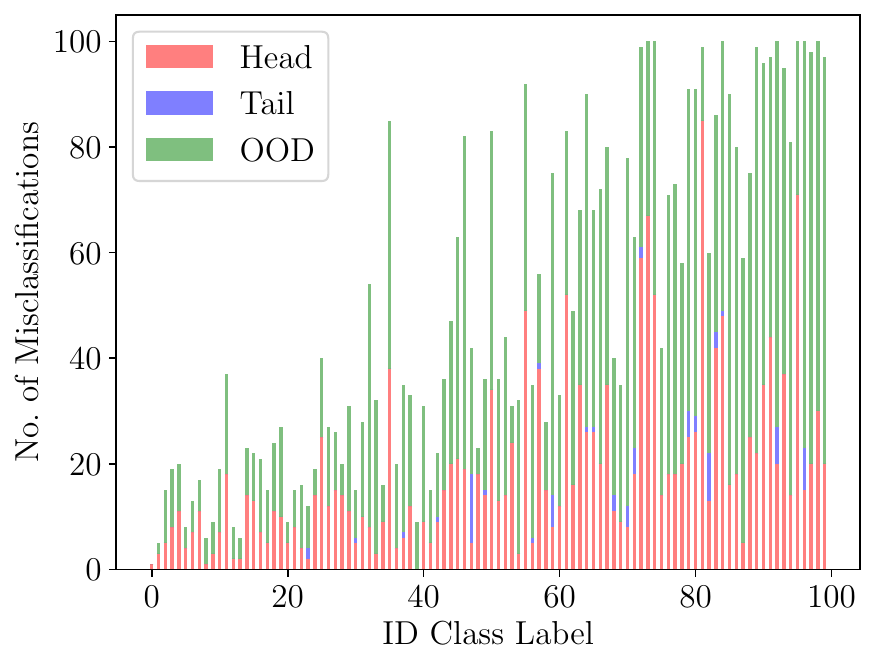}
    \caption{Misclassification Distribution - Baseline.}
    \label{fig:cifar100_misses_base}
  \end{subfigure}
    \hfill
  \begin{subfigure}{0.32\linewidth}
    \includegraphics[width=\textwidth]{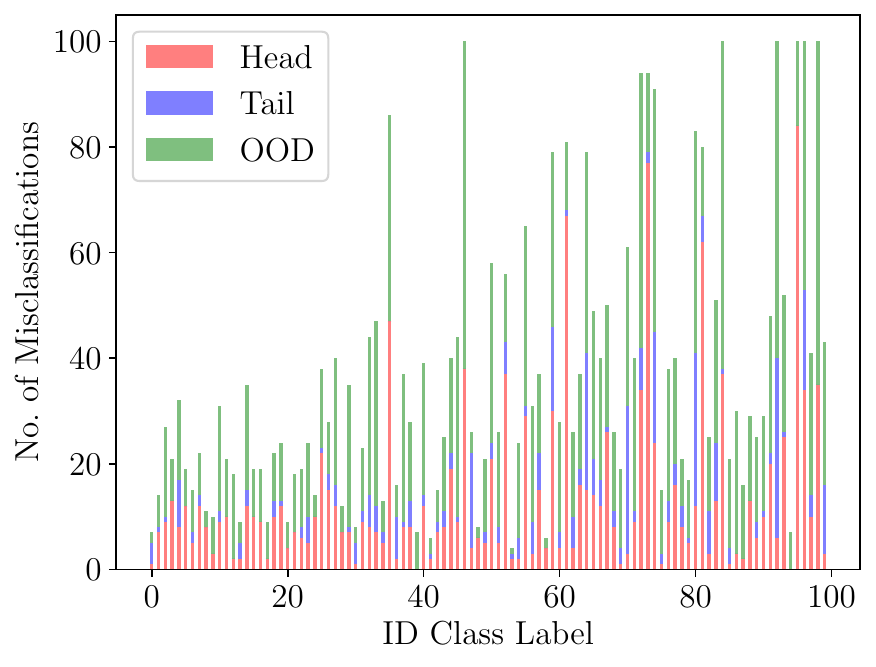}
    \caption{Misclassification Distribution - Ours.}
    \label{fig:cifar100_misses_ours}
  \end{subfigure}
    \hfill
    \begin{subfigure}{0.32\linewidth}
    \includegraphics[width=\textwidth]{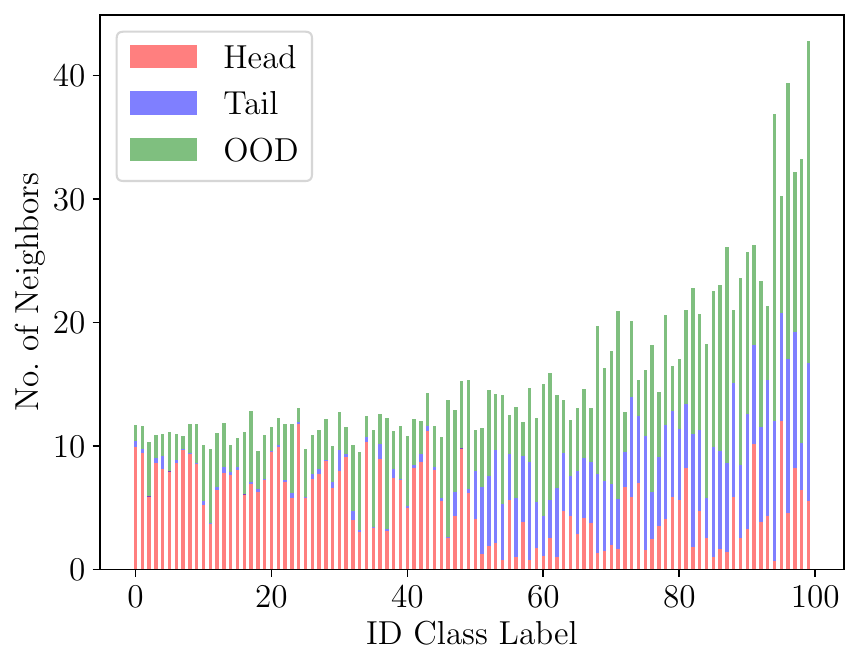}
    \caption{Neighbor Node Distribution.}
    \label{fig:cifar100_knn_neighbors}
  \end{subfigure}
  \vspace{-0.1in}
  \caption{Distribution of number of ID misclassifications over each class for CIFAR100-LT, with (a) one of the baselines ``Pre-train'' and (b) our method. (c) Head-class ID, tail-class ID and OOD neighbor node distribution over each ID class of CIFAR100-LT. Average neighbor values per each class are plotted due to the long-tailed nature. ID class labels higher than 50 belong to tail-classes.}
  \label{fig:cifar100_misclassification_analysis}
\vspace{-0.1in}
\end{figure*}

\subsection{Results analysis}
OOD detection evaluation results for CIFAR10-LT and CIFAR100-LT as IDs on individual \(\mathcal{D}_{\mathrm{out}}^{\mathrm{test}}\) datasets, are summarized in \cref{tab:cifar10_results} and \cref{tab:cifar100_results}, respectively. Overall results for both OOD detection and ID classification with additional baselines for all three ID datasets are summarized in \cref{tab:main_results}. For CIFAR10-LT and CIFAR100-LT, average values on all six datasets in the SC-OOD benchmark are reported. 

The outcomes of our study reveal a noteworthy superiority of our approach over other baseline methods in both OOD detection and ID classification across all ID datasets. Notably, in the more demanding CIFAR100-LT scenario, our method exhibits substantial improvements, having enhanced average values of 6.05\% for AUROC, 8.19\% for AUPR, 8.51\% for FPR95, and 6.95\% for ACC, respectively, compared to the second best method.

In the context of large-scale ImageNet-LT, which involves a broader spectrum of IDs, similar improvements of 2.49\% for AUROC and 1.53\% for AUPR are observed. The discernible trend in our findings indicates a pronounced enhancement in overall ID classification accuracy, particularly in the tail-classes. This is evidenced by the 10.64\% improvement in CIFAR10-LT, 12.95\% in CIFAR100-LT, and 1\% in ImageNet-LT, underscoring the substantial benefits our method brings to the tail-classes.


\subsection{Ablation study}
\noindent\textbf{Importance of different components:}
We present the results of ablation studies on each of the three key components outlined in our methodology in \cref{sec:method}: (i) graph representation learning (indicated as GRL), (ii) selection of pre-trained embeddings, and (iii) Gaussianization of pre-trained activation layers (indicated as Gau.).

To verify the performance gain from pre-training, we substitute the pre-trained backbone model with a baseline ResNet18 trained from scratch using OE~\cite{hendrycks_2018OE} to extract the initial features needed for the input graph. The results (denoted as Scratch+GCN in \cref{tab:ablation_results}) indicate a notable decrease compared to the results achieved with the Pre-train+GCN combination. Also, applying Gaussianization to pre-trained activation layers (Pre-train+Gau.+GCN) further improves the performance by a large margin compared to Pre-train+GCN alone. Furthermore, note that applying GRL on top of a backbone trained from scratch barely brings an improvement compared to a linear classifier on the same backbone (Scratch+GCN vs.~Scratch). This indicates the importance of selecting the initial feature representations of the graph for GRL to be successful.


\noindent\textbf{Importance of graph structure:}
To shed light on the importance of leveraging inter-sample relationships within the data, we also look at the distribution of misclassifications for CIFAR100-LT.
As shown in~\cref{fig:cifar100_misses_base}, the performance of the baseline model in the long-tailed setting gets worse as we move towards rare classes.
Interestingly, the OOD misses are also substantially higher for the tail classes.
These observations demonstrate the need for a better feature representation to simultaneously (i) distinguish the tail classes from the head, and (ii) separate the tail from the OOD data.
Our graph structure is designed to facilitate this.

In~\cref{fig:cifar100_knn_neighbors} we also plot the neighbor node distribution of input graph obtained by averaging the number of tail, head, and OOD neighbors for each ID node within a class.
As can be seen in this figure, the proposed graph structure creates a clear distinction between the head and tail nodes.
In particular, we see that most of the neighbors for the tail classes are also tail class.
Additionally, most OOD samples are assigned to a neighborhood of tail samples where there is more ambiguity that needs to be assessed.
The successful performance of our method, as shown in~\cref{fig:cifar100_misses_ours}, demonstrates that adding an inductive bias leveraging the inter-sample information between data points can help us protect the tail classes and have a more balanced error distribution.

More ablation studies on model architecture, graph structure, pre-training dataset, computational cost, and sensitivity analysis on key hyper-parameters can be found in the supplementary material.

\section{Discussion and conclusion}
Developing models robust to OOD samples in the presence of long-tailed IDs is important and challenging at the same time.
Specifically, it is vital to maintain accuracy for tail-class categories while ensuring robustness to OOD samples, particularly in certain applications.
For instance, consider the task of automated skin cancer detection. It is essential to distinguish between known rare skin lesion categories and unknown lesions effectively.

In this paper, we proposed a graph-based solution to address this problem on vision datasets. 
To this end, we used the $k$-NN graph built over an appropriate initial representation. By leveraging the feature space of a pre-trained model, which is adjusted using the process of Gaussianization, our graph structure can fully exploit the complex inter-relationships within the data.
We further refined our initial graph structure using GCNs to arrive at a feature space suitable for our intended downstream task. 
Through extensive experiments on both OOD detection and ID classification, we demonstrate the effectiveness of our method establishing a new SOTA. 

\looseness=-1
\noindent\textbf{Acknowledgments.} This research was undertaken using the LIEF HPC-GPGPU Facility hosted at the University of Melbourne. This Facility was established with the assistance of LIEF Grant LE170100200.
Sarah Erfani is in part supported by Australian Research Council (ARC) Discovery Early Career Researcher Award (DECRA) DE220100680.

{\small
\bibliographystyle{ieee_fullname}
\bibliography{egbib}
}

 \clearpage
\appendix
\setcounter{page}{1}

\twocolumn[{%
 \centering
 \Large \textbf{Exploiting Inter-Sample Information for Long-tailed Out-of-Distribution Detection}\\[2.0em]
 \vspace{-0.15in}
 \large Supplementary Material\\[3em]
}]

\section{Hyper-parameter tuning}
We used a validation set of 10\% and 17\% of training data for CIFAR10/100-LT and ImageNet-LT respectively. For experiments on CIFAR10/100-LT, we empirically set \(k=7\) to create the \(k\)-NN graphs with self-loops using 512-dimensional pre-trained embeddings.
The pre-trained model is trained for 100 epochs using SGD with Nesterov momentum and a cosine learning rate, with batch size 256 following \cite{hendrycks_2019a_using_pre}.
To update the BN statistics, BN layers are fine-tuned for another 20 epochs with a learning rate of 0.001 using all the training data.
GCN is trained for 200 epochs using the Adam optimizer with an initial learning rate of 0.001 and a cosine annealing learning rate scheduler. For ImageNet-LT, \(k\) is set to 2 and pre-trained embeddings are 2048-dimensional. BN layers are fine-tuned for 20 epochs with a learning rate of 0.001. GCN is trained for 250 epochs using the Adam optimizer with an initial learning rate of 0.01 which is decayed by a factor of 10 at epoch 15 and 100. On all datasets, we set \(\lambda=0.5\) following \cite{pascl}.

\section{Additional ablation studies}
\label{sec:more_ablation}

\subsection{Sensitivity analysis on $k$ and $\lambda$}
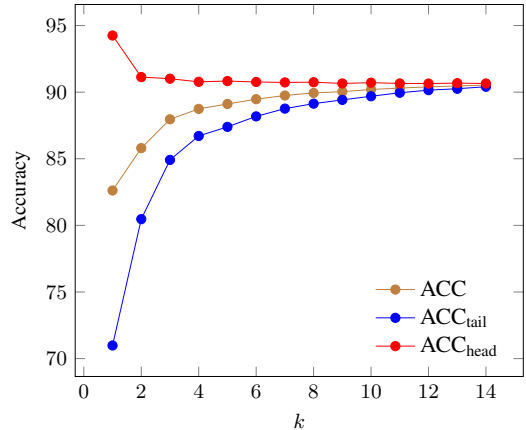
\begin{figure}[tb!]
\centering
\resizebox{0.4\textwidth}{!}{
\begin{tikzpicture}
\begin{axis}[
xlabel=$k$,
ylabel={Accuracy},
label style={font=\small},
tick label style={font=\small},
legend cell align={left},
clip=false,
legend pos=south east,
legend style={draw=none}
]
\addplot[color=brown, mark=*] table [x=k, y=ACC, col sep=comma] {all_k_pretrained_GCN_indctv_ACC_cifar10.csv};
\addlegendentry{ACC}
\addplot[color=blue, mark=*] table [x=k, y=tail, col sep=comma] {all_k_pretrained_GCN_indctv_ACC_cifar10.csv};
\addlegendentry{ACC\textsubscript{tail}}
\addplot[color=red, mark=*] table [x=k, y=head, col sep=comma] {all_k_pretrained_GCN_indctv_ACC_cifar10.csv};
\addlegendentry{ACC\textsubscript{head}}
\end{axis}
\end{tikzpicture}}
  \caption{\looseness=-1 Variation of CIFAR10-LT ID classification accuracy over selection of $k$ in the $k$-NN graph. }
  \label{fig:acc_plot}
\end{figure}

As seen, given a meaningful feature space, our graph combination can achieve a remarkable improvement, especially in terms of FPR95 and ACC\textsubscript{tail}. However, the improvement in tail-classes compensates for a slight decline in ACC\textsubscript{head}. This compensation can be attributed to the interplay between samples from tail-classes and the representative samples from head-classes. To provide a better explanation, we observe the behaviour of head and tail classes concerning the number of neighbor connections available for each sample. By increasing the $k$ in the $k$-NN graph, we increase the number of edges in the overall graph. As we can see in \cref{fig:acc_plot}, the increase in $k$ causes ACC\textsubscript{head} to drop, while ACC\textsubscript{tail} continues to grow along with the overall ACC. This suggests that too much neighborhood information around head-samples is undesirable, leading to over-smoothing of representative samples. In contrast, this additional information benefits the tail-samples, resulting in a much-improved ACC\textsubscript{tail} and a higher overall ACC.

We also conduct an ablation study on $\lambda$ that controls the importance of the OE loss term as summarized in~\cref{tab:lambda_ablation_results}. The performance of our method is stable with respect to different $\lambda$ values, showing a notable decrease only when $\lambda$ is set to 0.

\begin{table}[]
\centering
\caption{Ablation study on $\lambda$ for CIFAR10-LT as ID. Average values on all six \(\mathcal{D}_{\mathrm{out}}^{\mathrm{test}}\) are reported.}
\setlength\tabcolsep{0.45em}
\begin{small}
\begin{tabular}{c|ccc|ccc}
\hline
\textbf{$\lambda$} & \textbf{AUROC} & \textbf{AUPR} & \textbf{FPR95} & \textbf{ACC} & \textbf{ACC\textsubscript{head}} & \textbf{ACC\textsubscript{tail}} \\ \hline
0.00            & 96.95          & 96.58         & 12.10          & 89.77        & 90.84            & 88.70            \\
0.25            & 97.33          & 97.03         & 10.70          & 89.77        & 90.78            & 88.76            \\
0.50            & 97.42          & 97.15         & 10.58          & 89.73        & 90.74            & 88.73            \\
1.00            & 97.46          & 97.22         & 10.70          & 89.64        & 90.70            & 88.58            \\ \hline
\end{tabular}
\end{small}
\label{tab:lambda_ablation_results}
\end{table}

\begin{table*}[tb!]
\centering
\caption{Ablation study on model architecture for CIFAR10-LT and CIFAR100-LT as ID. Average values on all six \(\mathcal{D}_{\mathrm{out}}^{\mathrm{test}}\) are reported. The best results are shown in bold. Mean over six random runs are reported.}
\vspace{-0.10in}
\begin{small}
\begin{tabular}{c|c|ccc|ccc}
\hline
\textbf{$\mathcal{D}_{\mathrm{in}}$}                 & \textbf{Method} & \textbf{AUROC} & \textbf{AUPR} & \textbf{FPR95} & \textbf{ACC} & \textbf{ACC}\textsubscript{\textbf{head}} & \textbf{ACC}\textsubscript{\textbf{tail}} \\ \hline
\multirow{4}{*}{CIFAR10-LT}  & ResNet18        & 92.59          & 93.15          & 34.06          & 80.23          & 94.18            & 66.28            \\
                             & ResNet18+GCN    & 97.43          & {97.17} & 10.60          & 89.74          & 90.71            & 88.76            \\ \cline{2-8} 
                             & ViT-L\_16       & 97.81          & 96.55          & 7.66           & 94.97          & {97.30}   & 92.64            \\
                             & ViT-L\_16+GCN   & {98.34} & 97.10          & {4.36}  & {95.80} & 96.18            & {95.42}   \\ \hline
\multirow{4}{*}{CIFAR100-LT} & ResNet18        & 78.01          & 74.64          & 64.11          & 55.66          & 77.94            & 33.38            \\
                             & ResNet18+GCN    & 85.09          & {82.19} & 50.34          & 62.97          & 74.76            & 51.17            \\ \cline{2-8} 
                             & ViT-L\_16       & 86.81          & 77.51          & 37.51          & 69.74          & {88.62}   & 50.86            \\
                             & ViT-L\_16+GCN   & {89.26} & 79.33          & {28.31} & {74.90} & 86.20            & {63.61}   \\ \hline
\end{tabular}
\end{small}
\label{tab:ablation_model_archi}
\end{table*}

\begin{table*}[tb!]
\centering
\caption{Ablation study on batch-wise inference for CIFAR10-LT as ID. Average values on all six \(\mathcal{D}_{\mathrm{out}}^{\mathrm{test}}\) are reported. The best results are shown in bold. Mean over six random runs are reported.}
\vspace{-0.1in}
\begin{small}
\begin{tabular}{c|c|c|c|c|c|c}
\hline
\textbf{Batch   Size} & \textbf{AUROC} & \textbf{AUPR}  & \textbf{FPR95} & \textbf{ACC}   & \textbf{ACC}\textsubscript{\textbf{head}} & \textbf{ACC}\textsubscript{\textbf{tail}} \\ \hline
512                   & 95.96          & 95.84          & 16.43          & 83.19          & 87.87             & 78.50             \\
1024                  & 96.57          & 96.41          & 14.13          & 86.35          & 88.81             & 83.90             \\
2048                  & 96.86          & 96.66          & 13.01          & 87.58          & 89.37             & 85.80             \\
4096                  & 97.11          & 96.90          & 11.98          & 88.63          & 90.14             & 87.11             \\ 
8192                  & 97.25          & 96.97          & 11.29          & 89.15          & 90.40             & 87.90             \\ \hline
All Samples           & \textbf{97.43} & \textbf{97.17} & \textbf{10.60} & \textbf{89.74} & \textbf{90.71}    & \textbf{88.76}    \\ \hline
\end{tabular}
\end{small}
\label{tab:ablation_single_sample_cifar10}
\end{table*}

\begin{table*}[tb!]
\centering
\caption{Ablation study on batch-wise inference for CIFAR100-LT as ID. Average values on all six \(\mathcal{D}_{\mathrm{out}}^{\mathrm{test}}\) are reported. The best results are shown in bold. Mean over six random runs are reported.}
\vspace{-0.1in}
\begin{small}
\begin{tabular}{c|c|c|c|c|c|c}
\hline
\textbf{Batch   Size} & \textbf{AUROC} & \textbf{AUPR}  & \textbf{FPR95} & \textbf{ACC}   & \textbf{ACC}\textsubscript{\textbf{head}} & \textbf{ACC}\textsubscript{\textbf{tail}} \\ \hline
512                 & 69.15          & 67.49          & 77.66          & 32.90          & 48.13             & 17.68             \\ 
1024                & 75.22          & 73.56          & 69.90          & 42.17          & 57.75             & 26.59             \\ 
2048                & 79.81          & 77.79          & 62.14          & 50.45          & 65.33             & 35.56             \\ 
4096                & 82.51          & 80.08          & 56.43          & 56.09          & 69.73             & 42.46             \\ 
8192                & 83.84          & 81.21          & 53.30          & 59.42          & 71.94             & 46.90             \\ \hline
All Samples         & \textbf{85.09} & \textbf{82.19} & \textbf{50.34} & \textbf{62.97} &  \textbf{74.76}   & \textbf{51.17}    \\ \hline
\end{tabular}
\end{small}
\label{tab:ablation_single_sample_cifar100}
\end{table*}

\begin{figure*}[tb!]
  \centering
    \begin{subfigure}{1.0\linewidth}
    \includegraphics[width=\textwidth]{Images/legend.pdf}
    \caption*{}
    \label{fig:legend2}
  \end{subfigure}
      \\
  \begin{subfigure}{0.35\linewidth}
    \includegraphics[width=\textwidth]{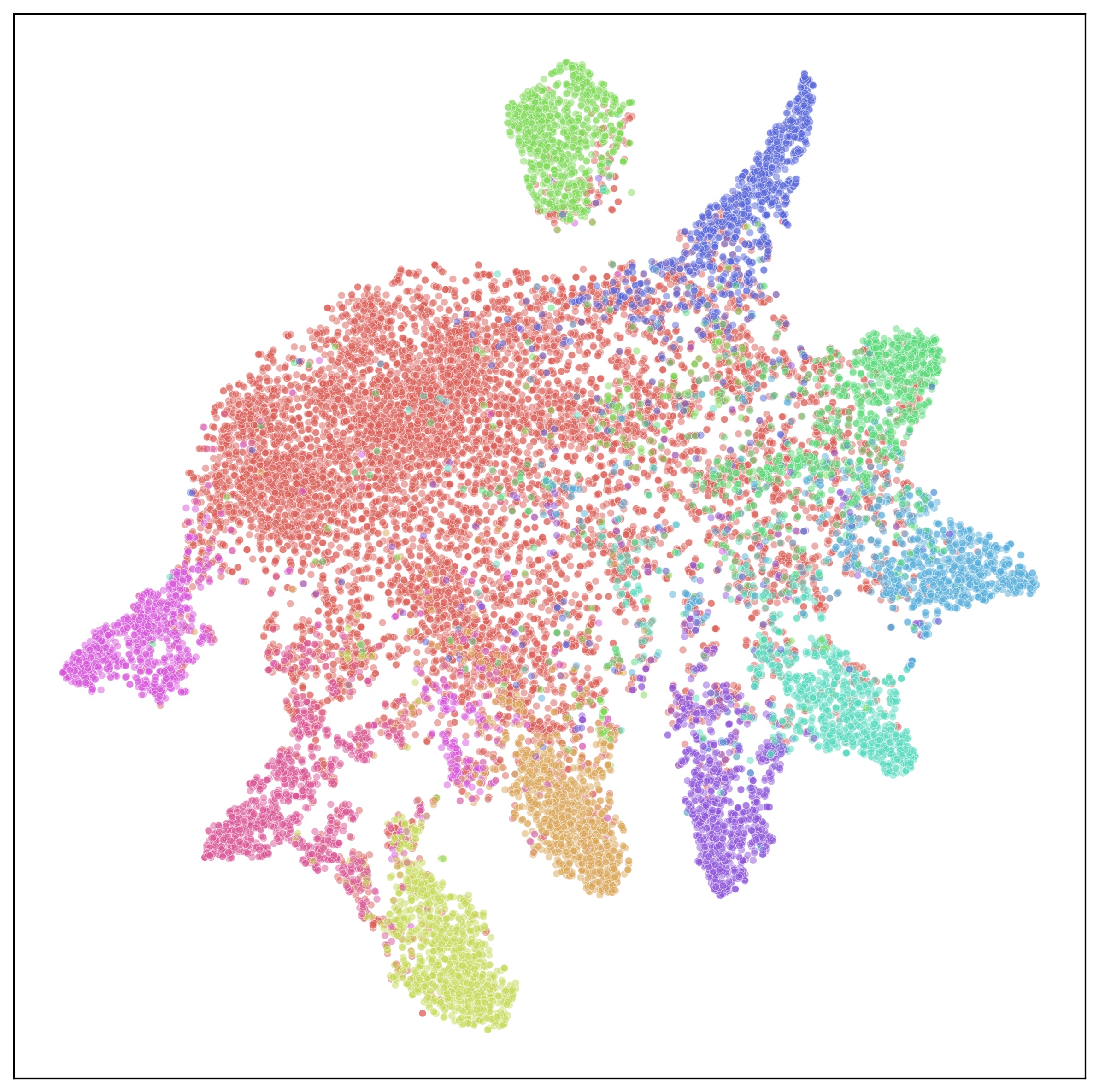}
    \caption{MLP.}
    \label{fig:MLP-a}
  \end{subfigure}
  \hspace{2em}
  \begin{subfigure}{0.35\linewidth}
    \includegraphics[width=\textwidth]{Images/new2_tsne_latent_GCN_cifar10cifar100.jpeg}
    \caption{GCN.}
    \label{fig:GCN-b}
  \end{subfigure}
  \caption{Feature space representations obtained from the (a) MLP and (b) GCN models for CIFAR10-LT as ID test set and CIFAR-100 as OOD test set using t-SNE. Equal number of test samples from each ID class are visualized. The GCN demonstrates distinct gaps between the decision boundaries of ID and OOD samples compared to the MLP.
  }
  \label{fig:feauture_space_vth_MLP}
\end{figure*}

\subsection{Model architecture}
In \cref{tab:main_results}, we employ ResNet18 as the pre-trained backbone model, aligning with prior work~\cite{pascl,choiBalancedEnOE} that addresses OOD detection in long-tailed recognition. In this section, we demonstrate the versatility of our method across various model architectures. Specifically, Fort \etal~\cite{VIT_exploringlimitsOOD} suggested that utilizing a pre-trained ViT can enhance OOD detection, although they did not evaluate their method on long-tailed IDs. We therefore replace pre-trained ResNet18 model with a ViT-L\_16 model pre-trained on ImageNet-1k~\cite{ImageNet} dataset. We observed that for ViTs, it would be more helpful to fine-tune the entire network before feature extraction. This results in even better performance (ViT-L\_16+GCN in \cref{tab:ablation_model_archi}) than the ResNet18 (ResNet18+GCN) underscoring the effectiveness of our method across diverse model architectures.

\subsection{An alternative inference method}
In our method, during inference, we construct a test graph using all ID and OOD test samples. 
However, in practical scenarios, inference is typically performed on batches of test samples which may belong to either ID or OOD. 
To ensure compatibility with such practical scenarios, we conduct experiments by processing batches of test samples instead of the full test data. 
In particular, for each randomly selected batch of data, we generate the $k$-NN graph over these data points alone to make the inference on that specific batch. 
Results for CIFAR10-LT and CIFAR100-LT as ID are indicated in \cref{tab:ablation_single_sample_cifar10} and \cref{tab:ablation_single_sample_cifar100}, respectively. 
Results indicate that our method exhibits stability in batch inference, delivering nearly comparable performance to processing all samples, particularly noticeable with larger batch sizes.

\subsection{Effect of the pre-training dataset}
To further examine the effect of the pre-training dataset, similar to an ablation study of~\cite{hendrycks_2019a_using_pre}, we removed 153 CIFAR-10 related classes from Downsampled ImageNet and reran the experiments for CIFAR10-LT. This further ensures the pre-trained model has not seen any tail-class or super-class related instances. Results shown in~\cref{tab:cifar_exclude} indicate similar performance gains. As we can see, our method still achieves SOTA (\wrt other baselines in~\cref{tab:main_results}) with only a small degradation compared to the model that uses all samples. This demonstrates that the
pre-trained model do not rely on seeing CIFAR-10 related samples, and that simply training on more natural images increases the overall performance~\cite{hendrycks_2019a_using_pre}.

\begin{table}[]
\caption{Performance on CIFAR10-LT with all CIFAR10-related classes removed from the pre-training dataset. Average values on all six \(\mathcal{D}_{\mathrm{out}}^{\mathrm{test}}\) are reported.}
\begin{small}
\setlength{\tabcolsep}{2.0pt} 
\begin{tabular}{c|ccc|ccc}
\hline
\textbf{Method} & \textbf{AUROC} & \textbf{AUPR} & \textbf{FPR95} & \textbf{ACC} & \textbf{ACC}\textsubscript{\textbf{head}} & \textbf{ACC}\textsubscript{\textbf{tail}} \\ \hline
cifar\_excluded & 96.55          & 96.33         & 14.66          & 87.78        & 88.95             & 86.60             \\
All             & 97.43          & 97.17         & 10.60          & 89.74        & 90.71             & 88.76             \\ \hline
\end{tabular}
\end{small}
\label{tab:cifar_exclude}
\end{table}

\subsection{Graph structure}
To further validate the importance of the graph structure, we replace the shallow linear classifier in the ``Pre-train+Gau.'' baseline with a deeper Multi-Layer Perceptron~(MLP) classifier. This MLP has an equivalent number of layers and trainable parameters as the GCN. Results for CIFAR10-LT and CIFAR100-LT as ID are shown in \cref{tab:ablation_MLP_cifar10} and \cref{tab:ablation_MLP_cifar100}, respectively. As the MLP does not consider any inter-sample relationships, the outcome demonstrates lower performance for the ``Pre-train+Gau.+MLP'' baseline compared to the combination with GCN (Pre-train+Gau.+GCN).
This observation is particularly evident in terms of FPR95 and ACC\textsubscript{tail}.

This fact can be further clarified by examining the feature space representations from these two models, as depicted in \cref{fig:feauture_space_vth_MLP}. 
As seen, there are distinct gaps between the decision boundaries of ID and OOD samples in the representations obtained from the GCN (\cref{fig:GCN-b}) compared to the MLP (\cref{fig:MLP-a}). 

\begin{table}[tb!]
\centering
\caption{Ablation study on importance of GRL using a MLP for CIFAR10-LT as ID. The best results are shown in bold. Mean over six random runs are reported.}
\vspace{-0.10in}
\hfill
\begin{subtable}{0.49\textwidth}
\centering
\caption{OOD detection results.}
\begin{footnotesize}
\setlength{\tabcolsep}{3pt} 
\renewcommand{\arraystretch}{1.1} 
\begin{tabular}{c|c|ccc}
\hline
\textbf{$\mathcal{D}_{\mathrm{out}}^{\mathrm{test}}$}           & \textbf{Method}      & \textbf{AUROC} & \textbf{AUPR}  & \textbf{FPR95} \\ \hline
\multirow{2}{*}{Texture}      & Pre-train+Gau.+MLP & 99.12          & 98.54          & 3.21           \\ \cline{2-5} 
                              & Pre-train+Gau.+GCN & \textbf{99.50} & \textbf{98.92} & \textbf{1.10}  \\ \hline
\multirow{2}{*}{SVHN}         & Pre-train+Gau.+MLP & 99.35          & \textbf{99.73} & 2.61           \\ \cline{2-5} 
                              & Pre-train+Gau.+GCN & \textbf{99.68} & 99.71          & \textbf{0.67}  \\ \hline
\multirow{2}{*}{CIFAR100}     & Pre-train+Gau.+MLP & 91.11          & 91.70          & 41.16          \\ \cline{2-5} 
                              & Pre-train+Gau.+GCN & \textbf{92.08} & \textbf{92.21} & \textbf{35.28} \\ \hline
\multirow{2}{*}{TinyImageNet} & Pre-train+Gau.+MLP & 94.40          & 92.93          & 28.66          \\ \cline{2-5} 
                              & Pre-train+Gau.+GCN & \textbf{95.61} & \textbf{93.64} & \textbf{19.93} \\ \hline
\multirow{2}{*}{LSUN}         & Pre-train+Gau.+MLP & 98.85          & 98.84          & 4.87           \\ \cline{2-5} 
                              & Pre-train+Gau.+GCN & \textbf{99.57} & \textbf{99.43} & \textbf{0.74}  \\ \hline
\multirow{2}{*}{Places365}    & Pre-train+Gau.+MLP & 97.55          & \textbf{99.17} & 14.86          \\ \cline{2-5} 
                              & Pre-train+Gau.+GCN & \textbf{98.12} & 99.12          & \textbf{5.89}  \\ \hline
\multirow{2}{*}{Average}      & Pre-train+Gau.+MLP & 96.73          & 96.82          & 15.90          \\ \cline{2-5} 
                              & Pre-train+Gau.+GCN & \textbf{97.43} & \textbf{97.17} & \textbf{10.60} \\ \hline
\end{tabular}
\end{footnotesize}
\label{tab:ablation_MLP_cifar10_OOD}
\end{subtable}
\hfill
\begin{subtable}{0.49\textwidth}
\centering
\caption{ID classification results.}
\begin{footnotesize}
\renewcommand{\arraystretch}{1.1} 
\begin{tabular}{c|c|c|c}
\hline
\textbf{Method}      & \textbf{ACC}   & \textbf{ACC}\textsubscript{\textbf{head}} & \textbf{ACC}\textsubscript{\textbf{tail}} \\ \hline
Pre-train+Gau.+MLP & 87.63          & \textbf{94.64}    & 80.61               \\ \hline
Pre-train+Gau.+GCN & \textbf{89.74} & 90.71             & \textbf{88.76}      \\ \hline
\end{tabular}
\end{footnotesize}
\label{tab:ablation_MLP_cifar10ACC}
\end{subtable}
\hfill
\label{tab:ablation_MLP_cifar10}
\end{table}

\begin{table}[tb!]
\centering
\caption{Ablation study on importance of GRL using a MLP for CIFAR100-LT as ID. The best results are shown in bold. Mean over six random runs are reported.}
\vspace{-0.1in}
\hfill
\begin{subtable}{0.49\textwidth}
\centering
\caption{OOD detection results.}
\begin{footnotesize}
\setlength{\tabcolsep}{3pt} 
\renewcommand{\arraystretch}{1.1} 
\begin{tabular}{c|c|ccc}
\hline
\textbf{$\mathcal{D}_{\mathrm{out}}^{\mathrm{test}}$}           & \textbf{Method}      & \textbf{AUROC} & \textbf{AUPR}  & \textbf{FPR95} \\ \hline
\multirow{2}{*}{Texture}      & Pre-train+Gau.+MLP & \textbf{90.95} & \textbf{84.76} & 39.54          \\ \cline{2-5} 
                              & Pre-train+Gau.+GCN & 90.88          & 84.10          & \textbf{39.32} \\ \hline
\multirow{2}{*}{SVHN}         & Pre-train+Gau.+MLP & 95.20          & 97.51          & 18.89          \\ \cline{2-5} 
                              & Pre-train+Gau.+GCN & \textbf{97.02} & \textbf{98.19} & \textbf{10.19} \\ \hline
\multirow{2}{*}{CIFAR10}      & Pre-train+Gau.+MLP & 73.39          & 68.21          & 70.79          \\ \cline{2-5} 
                              & Pre-train+Gau.+GCN & \textbf{75.08} & \textbf{74.17} & \textbf{70.62} \\ \hline
\multirow{2}{*}{TinyImageNet} & Pre-train+Gau.+MLP & 79.19          & 67.42          & 65.46          \\ \cline{2-5} 
                              & Pre-train+Gau.+GCN & \textbf{80.06} & \textbf{68.96} & \textbf{63.48} \\ \hline
\multirow{2}{*}{LSUN}         & Pre-train+Gau.+MLP & 81.03          & 72.32          & 68.22          \\ \cline{2-5} 
                              & Pre-train+Gau.+GCN & \textbf{83.29} & \textbf{75.09} & \textbf{59.57} \\ \hline
\multirow{2}{*}{Places365}    & Pre-train+Gau.+MLP & 83.21          & 92.05          & 60.41          \\ \cline{2-5} 
                              & Pre-train+Gau.+GCN & \textbf{84.22} & \textbf{92.63} & \textbf{58.87} \\ \hline
\multirow{2}{*}{Average}      & Pre-train+Gau.+MLP & 83.83          & 80.38          & 53.88          \\ \cline{2-5} 
                              & Pre-train+Gau.+GCN & \textbf{85.09} & \textbf{82.19} & \textbf{50.34} \\ \hline
\end{tabular}
\end{footnotesize}
\label{tab:ablation_MLP_cifar100_OOD}
\end{subtable}
\hfill
\begin{subtable}{0.49\textwidth}
\centering
\caption{ID classification results.}
\begin{footnotesize}
\renewcommand{\arraystretch}{1.1} 
\begin{tabular}{c|c|c|c}
\hline
\textbf{Method}      & \textbf{ACC}   & \textbf{ACC}\textsubscript{\textbf{head}} & \textbf{ACC}\textsubscript{\textbf{tail}} \\ \hline
Pre-train+Gau.+MLP & 61.62          & \textbf{79.12}    & 44.13               \\ \hline
Pre-train+Gau.+GCN & \textbf{62.97} & 74.76             & \textbf{51.17}      \\ \hline
\end{tabular}
\end{footnotesize}
\label{tab:ablation_MLP_cifar100ACC}
\end{subtable}
\hfill
\label{tab:ablation_MLP_cifar100}
\end{table}

\subsection{Computational cost}
\label{sec:computational_cost}
\cref{tab:time_space} compares the training and inference times of our method with other baselines. For our method, the total training time includes the time taken for Gaussianization, graph creation, and GCN training. Despite these components, the overall training time remains shorter because we employed a lightweight GCN model with only 578,860 parameters.
This is in contrast to most of the other methods, which require 11,599,752 parameters to be trained from scratch.
However, similar to KNN method, inference takes slightly longer than other methods due to the overhead introduced by $k$ nearest neighbour search and the graph creation process. Notably, varying $k$ values does not significantly impact the inference speed.

\begin{table}[]
\centering
\caption{Comparison of the training and inference speed (averaged per-image) for CIFAR100-LT \vs CIFAR10 experiments on an NVIDIA-A100 GPU. For our method, different $k$ values (indicated in brackets) are compared.}
\begin{small}
\begin{tabular}{c|c|c}
\hline
\textbf{Method}                & \textbf{Train. time(s)} & \textbf{Infer. time(ms)}   \\ \hline
OECC                  & 715.57         & 0.27                               \\
EnergyOE              & 687.40         & 0.24                                 \\
OE                    & 690.09         & 0.11                                \\
PASCL                 & 1265.63        & 0.13                                \\
BE-OE                 & 706.87         & 0.24                               \\
EAT                   & 3526.58        & 0.14                                \\
COCL                  & 659.74         & 0.11                                 \\
Pre-train+KNN         & 603.96         & 1.67                                       \\ \hline
\multirow{3}{*}{Ours} & 478.51 (k=3)   & 2.09 (k=3)            \\
                      & 533.13 (k=7)   & 2.10 (k=7)                                    \\
                      & 620.10 (k=10)  & 2.11 (k=10)                                  \\ \hline
\end{tabular}
\end{small}
\label{tab:time_space}
\end{table}

\subsection{Performance on balanced ID datasets}
We also compare the performance of our method on balanced ID datasets in \cref{tab:balanced_ID}. Note that some LTR baselines, including PASCL, OS, EAT, and COCL (which we did not include in the table), cannot be directly evaluated or reduced to other baselines when there are zero tail-classes. For our method, we used the same experimental settings as in the LT experiments, except that the imbalance ratio $\rho$ was set to 1.
As shown, our method outperforms most baselines for OOD detection across all metrics while demonstrating comparable performance in ID classification accuracy.
There is a small effect of reduction of ID accuracy, when using higher $k$ values, which result in larger number of neighbors around ID nodes in the graph. 
However, the inter-sample relationships between nodes greatly benefit the OOD detection.
This is part of the trade-off between leveraging useful relational information and the risk of over-smoothing, a common issue inherent to many graph-based approaches.

\begin{table}[]
\caption{OOD detection results and ID classification results for balanced CIFAR-10 and CIFAR-100 as ID. Average values on all six \(\mathcal{D}_{\mathrm{out}}^{\mathrm{test}}\) are reported. The best results are bolded. Mean over six random runs are reported.}
\begin{small}
\setlength{\tabcolsep}{2.0pt} 
\begin{tabular}{c|ccccc}
\hline
\textbf{$\mathcal{D}_{\mathrm{in}}$}    & \textbf{Method} & \textbf{AUROC} & \textbf{AUPR}  & \textbf{FPR95} & \textbf{ACC}   \\ \hline
\multirow{6}{*}{CIFAR-10}  & OECC            & 96.33          & 95.38          & 14.36          & 91.57          \\
                             & EnergyOE        & 96.77          & 96.72          & 14.82          & 93.30          \\
                             & OE              & 96.67          & 95.97          & 13.80          & 93.64          \\
                             & BE-OE           & 96.83          & 96.70          & 14.51          & 93.00          \\
                             & Pre-train+KNN             & 94.82          & 94.03          & 26.36          & \textbf{95.91} \\
                             & Ours            & \textbf{97.84} & \textbf{97.66} & \textbf{8.86}  & 91.40          \\ \hline
\multirow{6}{*}{CIFAR-100} & OECC            & 84.03          & 77.94          & 45.26          & 69.55          \\
                             & EnergyOE        & 85.84          & 80.99          & 43.02          & 74.95          \\
                             & OE              & 83.92          & 78.07          & 49.49          & 71.36          \\
                             & BE-OE           & 85.85          & 80.91          & \textbf{42.93} & 74.83          \\
                             & Pre-train+KNN             & 85.52          & 80.71          & 55.56          & \textbf{79.82} \\
                             & Ours            & \textbf{87.17} & \textbf{83.91} & 43.62          & 70.37          \\ \hline
\end{tabular}
\end{small}
\label{tab:balanced_ID}
\end{table}

\subsection{Detailed ablation results}
In~\Cref{tab:ablation_results}, we presented an ablation study over various components of our proposed approach. 
Due to space limitations, we omitted the detailed results over separate OOD data.
Detailed ablation study results over separate \(\mathcal{D}_{\mathrm{out}}^{\mathrm{test}}\) for CIFAR10-LT and CIFAR100-LT are shown in \cref{tab:ablation_cifar10} and \cref{tab:ablation_cifar100}, respectively.

\section{More details on Gaussianization}
Regularization and normalization techniques are commonly applied to neural networks to ensure stable training and improved generalization performance~\cite{jooBN, BN}. Among various normalization methods for controlling hidden activations, Gaussianization ensures that the activation layer representations follow a standard normal distribution. This approach relies on batch normalization (BN) layers, which take hidden activations as input to the next layer and normalize them to have zero mean and unit variance. During inference, the network uses `running statistics' (running mean-\(\mu_{\text {running }}\) and running variance-\(\sigma_{\text {running }}\))  that are computed and updated during training to normalize activations, i.e.,
\begin{align}\nonumber
\hat{x}_i=\frac{x_i-\mu_{\text {running }}}{\sqrt{\sigma_{\text {running }}^2+\epsilon}},
\label{eq:bn} 
\end{align}
where \(x_i\) denotes the current data input and \(\epsilon\) is a small constant added for numerical stability.
In our pipeline, we used a backbone model \(f\) pre-trained on a different dataset ($\mathcal{D}_{\mathrm{pretrain}}$) to extract initial feature representations for each training input. However, when calculating these feature representations, the running statistics from $\mathcal{D}_{\mathrm{pretrain}}$ are used, which differ from those of the current training data ($\mathcal{D}_{\mathrm{train}}$).
This results in a misalignment of the activations from the intended standard normal distribution during inference on $\mathcal{D}_{\mathrm{train}}$.
To correct this, we need to fine-tune the BN layers of the backbone \(f\) on our training data $\mathcal{D}_{\mathrm{train}}$, while keeping all other layers fixed. This fine-tuning will update the \(\mu_{\text {running }}\) and \(\sigma_{\text {running }}\) to reflect the distribution of $\mathcal{D}_{\mathrm{train}}$, aligning the activations with the intended standard normalization.

\section{Limitations}
A common limitation of our method can be that computational overhead introduced by graph creation (\cref{sec:computational_cost}). Future work could explore other graph-based approaches, such as GraphSAGE~\cite{GraphSAGE}, which use neighborhood sampling to reduce computational costs.
On the other hand, there is a trade-off between leveraging useful relational information by means of message passing and the risk of over-smoothing, which is a common issue in many graph-based approaches. 
We leave it to future work to introduce edge-weighted mechanisms and graph sampling techniques to address these potential limitations.

\section{Code}
\label{sec:code}
Code to reproduce the results of our method is available at \url{https://github.com/hdnugit/Graph_OOD_LTR}.

\begin{table*}[p!]
\centering
\caption{Ablation study for separate \(\mathcal{D}_{\mathrm{out}}^{\mathrm{test}}\) for CIFAR10-LT as ID. The best and second-best results are bolded and underlined, respectively. Mean over six random runs are reported.}
\begin{small}
\begin{tabular}{c|c|c|c|c|c|c|c}
\hline
\textbf{$\mathcal{D}_{\mathrm{out}}^{\mathrm{test}}$}             & \textbf{\makecell{Pre-\\train}} & \textbf{Gau.} & \textbf{GRL} & \textbf{Method}               & \textbf{AUROC} & \textbf{AUPR}  & \textbf{FPR95} \\ \hline
\multirow{6}{*}{Texture}      &\xmark                     &\xmark               &\xmark            & Scratch (OE)                  & 92.59          & 83.32          & 25.10          \\ \cline{2-8} 
                              &\xmark                     &\xmark               &\cmark            & Scratch+GCN                   & 86.34          & 69.51          & 32.64          \\ \cline{2-8} 
                              &\cmark                     &\xmark               &\xmark            & Pre-train                     & 94.42          & 90.58          & 26.77          \\ \cline{2-8} 
                              &\cmark                     &\xmark               &\cmark            & Pre-train+GCN                 & {\ul 99.16}    & {\ul 98.53}    & {\ul 2.10}     \\ \cline{2-8} 
                              &\cmark                     &\cmark               &\xmark            & Pre-train+Gau.              & 96.58          & 94.49          & 19.23          \\ \cline{2-8} 
                              &\cmark                     &\cmark               &\cmark            & Pre-train+Gau.+GCN   (ours) & \textbf{99.50} & \textbf{98.92} & \textbf{1.10}  \\ \hline
\multirow{6}{*}{SVHN}         &\xmark                     &\xmark               &\xmark            & Scratch (OE)                  & 95.10          & 97.14          & 16.15          \\ \cline{2-8} 
                              &\xmark                     &\xmark               &\cmark            & Scratch+GCN                   & 87.67          & 91.66          & 32.16          \\ \cline{2-8} 
                              &\cmark                     &\xmark               &\xmark            & Pre-train                     & 97.00          & 98.71          & 15.64          \\ \cline{2-8} 
                              &\cmark                     &\xmark               &\cmark            & Pre-train+GCN                 & {\ul 99.44}    & {\ul 99.63}    & {\ul 1.72}     \\ \cline{2-8} 
                              &\cmark                     &\cmark               &\xmark            & Pre-train+Gau.              & 96.17          & 98.30          & 18.54          \\ \cline{2-8} 
                              &\cmark                     &\cmark               &\cmark            & Pre-train+Gau.+GCN   (ours) & \textbf{99.68} & \textbf{99.71} & \textbf{0.67}  \\ \hline
\multirow{6}{*}{CIFAR100}     &\xmark                     &\xmark               &\xmark            & Scratch (OE)                  & 83.40          & 80.93          & 56.96          \\ \cline{2-8} 
                              &\xmark                     &\xmark               &\cmark            & Scratch+GCN                   & 79.55          & 73.12          & 56.27          \\ \cline{2-8} 
                              &\cmark                     &\xmark               &\xmark            & Pre-train                     & 84.13          & 83.90          & 57.24          \\ \cline{2-8} 
                              &\cmark                     &\xmark               &\cmark            & Pre-train+GCN                 & {\ul 88.87}    & {\ul 88.38}    & {\ul 44.18}    \\ \cline{2-8} 
                              &\cmark                     &\cmark               &\xmark            & Pre-train+Gau.              & 85.92          & 86.48          & 55.13          \\ \cline{2-8} 
                              &\cmark                     &\cmark               &\cmark            & Pre-train+Gau.+GCN   (ours) & \textbf{92.08} & \textbf{92.21} & \textbf{35.28} \\ \hline
\multirow{6}{*}{TinyImageNet} &\xmark                     &\xmark               &\xmark            & Scratch (OE)                  & 86.14          & 79.33          & 47.78          \\ \cline{2-8} 
                              &\xmark                     &\xmark               &\cmark            & Scratch+GCN                   & 81.75          & 70.13          & 44.80          \\ \cline{2-8} 
                              &\cmark                     &\xmark               &\xmark            & Pre-train                     & 88.21          & 85.00          & 46.27          \\ \cline{2-8} 
                              &\cmark                     &\xmark               &\cmark            & Pre-train+GCN                 & {\ul 93.38}    & {\ul 90.34}    & {\ul 26.22}    \\ \cline{2-8} 
                              &\cmark                     &\cmark               &\xmark            & Pre-train+Gau.              & 90.20          & 88.02          & 42.69          \\ \cline{2-8} 
                              &\cmark                     &\cmark               &\cmark            & Pre-train+Gau.+GCN   (ours) & \textbf{95.61} & \textbf{93.64} & \textbf{19.93} \\ \hline
\multirow{6}{*}{LSUN}         &\xmark                     &\xmark               &\xmark            & Scratch (OE)                  & 91.35          & 87.62          & 27.86          \\ \cline{2-8} 
                              &\xmark                     &\xmark               &\cmark            & Scratch+GCN                   & 86.08          & 79.82          & 33.54          \\ \cline{2-8} 
                              &\cmark                     &\xmark               &\xmark            & Pre-train                     & 93.26          & 92.88          & 31.92          \\ \cline{2-8} 
                              &\cmark                     &\xmark               &\cmark            & Pre-train+GCN                 & {\ul 99.38}    & {\ul 99.37}    & {\ul 1.88}     \\ \cline{2-8} 
                              &\cmark                     &\cmark               &\xmark            & Pre-train+Gau.              & 94.30          & 94.40          & 30.40          \\ \cline{2-8} 
                              &\cmark                     &\cmark               &\cmark            & Pre-train+Gau.+GCN   (ours) & \textbf{99.57} & \textbf{99.43} & \textbf{0.74}  \\ \hline
\multirow{6}{*}{Places365}    &\xmark                     &\xmark               &\xmark            & Scratch (OE)                  & 90.07          & 95.15          & 34.04          \\ \cline{2-8} 
                              &\xmark                     &\xmark               &\cmark            & Scratch+GCN                   & 84.09          & 90.98          & 35.92          \\ \cline{2-8} 
                              &\cmark                     &\xmark               &\xmark            & Pre-train                     & 91.18          & 96.66          & 40.32          \\ \cline{2-8} 
                              &\cmark                     &\xmark               &\cmark            & Pre-train+GCN                 & {\ul 97.76}    & {\ul 99.09}    & {\ul 9.55}     \\ \cline{2-8} 
                              &\cmark                     &\cmark               &\xmark            & Pre-train+Gau.              & 92.37          & 97.25          & 38.36          \\ \cline{2-8} 
                              &\cmark                     &\cmark               &\cmark            & Pre-train+Gau.+GCN   (ours) & \textbf{98.12} & \textbf{99.12} & \textbf{5.89}  \\ \hline
\multirow{6}{*}{Average}      &\xmark                     &\xmark               &\xmark            & Scratch (OE)                  & 89.77          & 87.25          & 34.65          \\ \cline{2-8} 
                              &\xmark                     &\xmark               &\cmark            & Scratch+GCN                   & 84.25          & 79.20          & 39.22          \\ \cline{2-8} 
                              &\cmark                     &\xmark               &\xmark            & Pre-train                     & 91.37          & 91.29          & 36.36          \\ \cline{2-8} 
                              &\cmark                     &\xmark               &\cmark            & Pre-train+GCN                 & {\ul 96.33}    & {\ul 95.89}    & {\ul 14.28}    \\ \cline{2-8} 
                              &\cmark                     &\cmark               &\xmark            & Pre-train+Gau.              & 92.59          & 93.15          & 34.06          \\ \cline{2-8} 
                              &\cmark                     &\cmark               &\cmark            & Pre-train+Gau.+GCN   (ours) & \textbf{97.43} & \textbf{97.17} & \textbf{10.60} \\ \hline
\end{tabular}
\end{small}
\label{tab:ablation_cifar10}
\end{table*}

\begin{table*}[p!]
\centering
\caption{Ablation study for separate \(\mathcal{D}_{\mathrm{out}}^{\mathrm{test}}\) for CIFAR100-LT as ID. The best and second-best results are bolded and underlined, respectively. Mean over six random runs are reported.}
\begin{small}
\begin{tabular}{c|c|c|c|c|c|c|c}
\hline
\textbf{$\mathcal{D}_{\mathrm{out}}^{\mathrm{test}}$}             & \textbf{\makecell{Pre-\\train}} & \textbf{Gau.} & \textbf{GRL} & \textbf{Method}               & \textbf{AUROC} & \textbf{AUPR}  & \textbf{FPR95} \\ \hline
\multirow{6}{*}{Texture}      &\xmark                     &\xmark               &\xmark            & Scratch (OE)                  & 76.71          & 58.79          & 68.28          \\ \cline{2-8} 
                              &\xmark                     &\xmark               &\cmark            & Scratch+GCN                   & 78.70          & 56.03          & 62.57          \\ \cline{2-8} 
                              &\cmark                     &\xmark               &\xmark            & Pre-train                     & 77.39          & 62.69          & 65.45          \\ \cline{2-8} 
                              &\cmark                     &\xmark               &\cmark            & Pre-train+GCN                 & {\ul 88.70}    & {\ul 81.86}    & {\ul 46.35}    \\ \cline{2-8} 
                              &\cmark                     &\cmark               &\xmark            & Pre-train+Gau.              & 83.79          & 72.87          & 54.47          \\ \cline{2-8} 
                              &\cmark                     &\cmark               &\cmark            & Pre-train+Gau.+GCN   (ours) & \textbf{90.88} & \textbf{84.10} & \textbf{39.32} \\ \hline
\multirow{6}{*}{SVHN}         &\xmark                     &\xmark               &\xmark            & Scratch (OE)                  & 77.61          & 86.82          & 58.04          \\ \cline{2-8} 
                              &\xmark                     &\xmark               &\cmark            & Scratch+GCN                   & 79.04          & 85.16          & 62.77          \\ \cline{2-8} 
                              &\cmark                     &\xmark               &\xmark            & Pre-train                     & 83.80          & 91.15          & 49.25          \\ \cline{2-8} 
                              &\cmark                     &\xmark               &\cmark            & Pre-train+GCN                 & 89.42          & 93.85          & 34.56          \\ \cline{2-8} 
                              &\cmark                     &\cmark               &\xmark            & Pre-train+Gau.              & {\ul 91.65}    & {\ul 95.90}    & {\ul 33.24}    \\ \cline{2-8} 
                              &\cmark                     &\cmark               &\cmark            & Pre-train+Gau.+GCN   (ours) & \textbf{97.02} & \textbf{98.19} & \textbf{10.19} \\ \hline
\multirow{6}{*}{CIFAR10}      &\xmark                     &\xmark               &\xmark            & Scratch (OE)                  & 62.23          & 57.57          & 80.64          \\ \cline{2-8} 
                              &\xmark                     &\xmark               &\cmark            & Scratch+GCN                   & 64.98          & 59.53          & 78.15          \\ \cline{2-8} 
                              &\cmark                     &\xmark               &\xmark            & Pre-train                     & 64.02          & 60.22          & 80.23          \\ \cline{2-8} 
                              &\cmark                     &\xmark               &\cmark            & Pre-train+GCN                 & {\ul 72.80}    & {\ul 68.96}    & {\ul 71.20}    \\ \cline{2-8} 
                              &\cmark                     &\cmark               &\xmark            & Pre-train+Gau.              & 68.99          & 66.00          & 77.65          \\ \cline{2-8} 
                              &\cmark                     &\cmark               &\cmark            & Pre-train+Gau.+GCN   (ours) & \textbf{75.08} & \textbf{74.17} & \textbf{70.62} \\ \hline
\multirow{6}{*}{TinyImageNet} &\xmark                     &\xmark               &\xmark            & Scratch (OE)                  & 68.04          & 51.66          & 76.66          \\ \cline{2-8} 
                              &\xmark                     &\xmark               &\cmark            & Scratch+GCN                   & 68.37          & 51.73          & 77.60          \\ \cline{2-8} 
                              &\cmark                     &\xmark               &\xmark            & Pre-train                     & 71.47          & 56.95          & 74.29          \\ \cline{2-8} 
                              &\cmark                     &\xmark               &\cmark            & Pre-train+GCN                 & {\ul 75.73}    & {\ul 62.25}    & {\ul 69.33}    \\ \cline{2-8} 
                              &\cmark                     &\cmark               &\xmark            & Pre-train+Gau.              & 74.70          & 61.74          & 70.48          \\ \cline{2-8} 
                              &\cmark                     &\cmark               &\cmark            & Pre-train+Gau.+GCN   (ours) & \textbf{80.06} & \textbf{68.96} & \textbf{63.48} \\ \hline
\multirow{6}{*}{LSUN}         &\xmark                     &\xmark               &\xmark            & Scratch (OE)                  & 77.10          & 61.42          & {\ul 63.98}    \\ \cline{2-8} 
                              &\xmark                     &\xmark               &\cmark            & Scratch+GCN                   & 76.26          & 57.55          & 65.52          \\ \cline{2-8} 
                              &\cmark                     &\xmark               &\xmark            & Pre-train                     & 65.18          & 53.89          & 81.17          \\ \cline{2-8} 
                              &\cmark                     &\xmark               &\cmark            & Pre-train+GCN                 & {\ul 79.14}    & {\ul 69.66}    & 69.21          \\ \cline{2-8} 
                              &\cmark                     &\cmark               &\xmark            & Pre-train+Gau.              & 72.06          & 62.26          & 77.96          \\ \cline{2-8} 
                              &\cmark                     &\cmark               &\cmark            & Pre-train+Gau.+GCN   (ours) & \textbf{83.29} & \textbf{75.09} & \textbf{59.57} \\ \hline
\multirow{6}{*}{Places365}    &\xmark                     &\xmark               &\xmark            & Scratch (OE)                  & 75.80          & 86.68          & 65.72          \\ \cline{2-8} 
                              &\xmark                     &\xmark               &\cmark            & Scratch+GCN                   & 76.26          & 85.78          & 64.88          \\ \cline{2-8} 
                              &\cmark                     &\xmark               &\xmark            & Pre-train                     & 71.18          & 85.69          & 77.00          \\ \cline{2-8} 
                              &\cmark                     &\xmark               &\cmark            & Pre-train+GCN                 & {\ul 82.57}    & {\ul 91.73}    & {\ul 62.65}    \\ \cline{2-8} 
                              &\cmark                     &\cmark               &\xmark            & Pre-train+Gau.              & 76.89          & 89.04          & 70.86          \\ \cline{2-8} 
                              &\cmark                     &\cmark               &\cmark            & Pre-train+Gau.+GCN   (ours) & \textbf{84.22} & \textbf{92.63} & \textbf{58.87} \\ \hline
\multirow{6}{*}{Average}      &\xmark                     &\xmark               &\xmark            & Scratch (OE)                  & 72.91          & 67.16          & 68.89          \\ \cline{2-8} 
                              &\xmark                     &\xmark               &\cmark            & Scratch+GCN                   & 73.93          & 65.97          & 68.58          \\ \cline{2-8} 
                              &\cmark                     &\xmark               &\xmark            & Pre-train                     & 72.17          & 68.43          & 71.23          \\ \cline{2-8} 
                              &\cmark                     &\xmark               &\cmark            & Pre-train+GCN                 & {\ul 81.40}    & {\ul 78.05}    & {\ul 58.88}    \\ \cline{2-8} 
                              &\cmark                     &\cmark               &\xmark            & Pre-train+Gau.              & 78.01          & 74.64          & 64.11          \\ \cline{2-8} 
                              &\cmark                     &\cmark               &\cmark            & Pre-train+Gau.+GCN   (ours) & \textbf{85.09} & \textbf{82.19} & \textbf{50.34} \\ \hline
\end{tabular}
\end{small}
\label{tab:ablation_cifar100}
\end{table*}
\clearpage

\end{document}